\documentclass[11pt]{article}

\usepackage[T1]{fontenc}
\usepackage{newtxtext,newtxmath}
\usepackage[margin=1in]{geometry}
\usepackage{microtype}
\usepackage{amsmath}
\usepackage{booktabs,multirow,makecell}
\usepackage{graphicx}
\usepackage[table]{xcolor}
\usepackage{enumitem}
\usepackage{caption,subcaption}
\usepackage[numbers,sort&compress]{natbib}
\usepackage{pgfplots}
\usepackage{xurl}
\pgfplotsset{compat=1.17}
\usepackage[pdfusetitle,colorlinks=true,linkcolor=blue!60!black,citecolor=blue!60!black,urlcolor=blue!60!black]{hyperref}
\usepackage[capitalize,noabbrev]{cleveref}

\newcommand{\piz}{\pi_{0}}
\newcommand{\pibase}{\pi_{\mathrm{base}}}
\newcommand{\Trl}{T_{\mathrm{RL}}}
\newcommand{\Tf}{T_{\mathrm{F}}}
\newcommand{\pisftf}{\pi^{\mathrm{F}}_{\mathrm{SFT}}}
\newcommand{\pisftk}{\pi^{\mathrm{K}}_{\mathrm{SFT}}}
\newcommand{\avgk}[1]{\mathrm{Avg@}#1}
\newcommand{\bestk}[1]{\mathrm{Best@}#1}

\title{Proximity over Strength in On-Policy Distillation\texorpdfstring{\\[2pt]\large}{: }%
  Evidence from Teacher Swaps in Code and Math}

\author{Zhiyuan Ma\\
\texttt{zhiyuanma0520@gmail.com}}
\hypersetup{pdfauthor={Zhiyuan Ma},pdfsubject={Preprint, version 1}}

\date{}

\begin{document}
\maketitle

\begin{abstract}
In on-policy distillation, a teacher scores the reasoning its student writes, including choices the teacher would rarely make itself. We compared two ways to choose that teacher. One uses the teacher's own benchmark score. The other uses the mean log probability it assigns to the starting student's reasoning tokens. In each domain we trained a fresh copy of the same student with each teacher. One teacher was the student's \emph{future self}, a copy of the starting model that had already improved through reinforcement learning. On 172 competitive programming problems, this teacher assigned the highest probability on average to the student's choices and raised its mean score from 7.10 to 13.55 out of 100. With either of two stronger teacher models, the student's mean score fell below 4. A separate Flash recipe first trained the student on demonstrations. Adding program execution rewards gave no clear gain over its best distillation checkpoint. We identified the ranking retrospectively on code and tested it prospectively on competition math. The future-self student finished 7.0 percentage points ahead of the student taught by a stronger 32B model.
\end{abstract}

\section{Introduction}
\label{sec:intro}

A student can learn to solve a problem by reading a teacher's worked solution, as in supervised fine-tuning (SFT) on reasoning traces \citep{guo2025deepseekr1}. In on-policy distillation (OPD), the student attempts the problem itself \citep{agarwal2024gkd}. At each token, the teacher sees what the student has written so far and assigns a probability to the student's next choice. Comparing that probability with the student's own gives feedback along the whole attempt, even when the finished program earns no credit \citep{lu2025opd}.

We compare two ways to choose a teacher before OPD. One ranks candidates by their own benchmark scores. The other ranks them by the mean log probability they assign to reasoning tokens sampled from the starting student. We examined the ranking retrospectively on code and pre-registered the prediction on math. The models in both domains reason for tens of thousands of tokens. One teacher is the student's own checkpoint after outcome-reward reinforcement learning (RL), which we call its \emph{future self}.

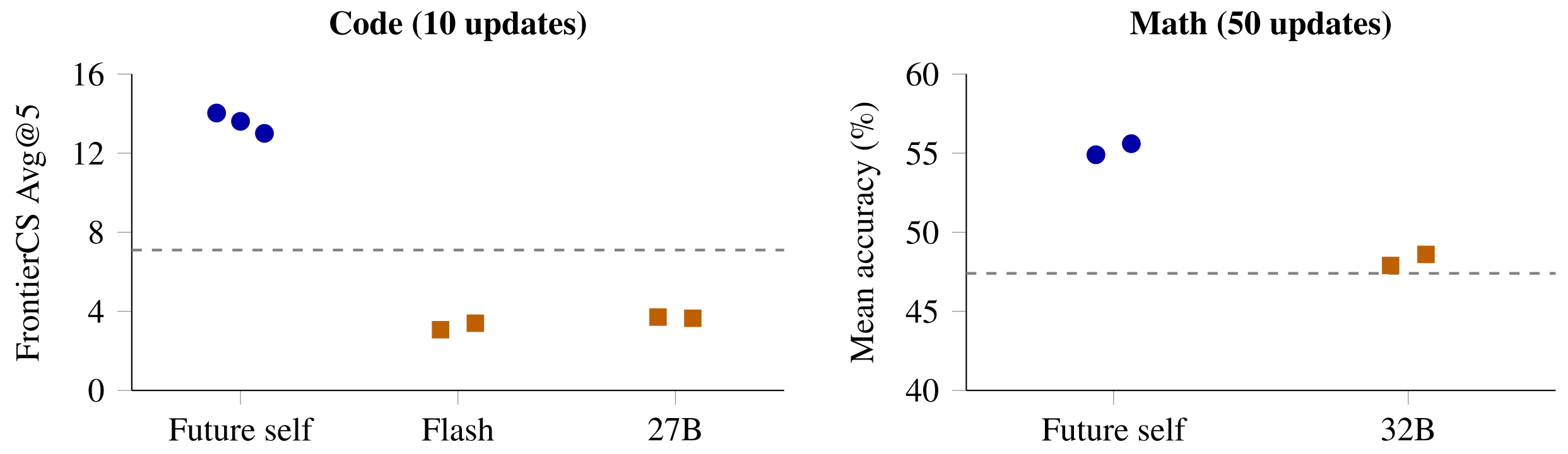
\begin{figure}[t]
\centering
\begin{minipage}[t]{0.51\linewidth}
\centering
\begin{tikzpicture}
\begin{axis}[
  width=\linewidth, height=4.9cm, title={Code (10 updates)},
  ylabel={FrontierCS $\avgk{5}$}, xmin=0.5, xmax=3.5, ymin=0, ymax=16,
  xtick={1,2,3}, xticklabels={Future self,Flash,27B}, ytick={0,4,8,12,16},
  tick label style={font=\small}, label style={font=\small},
  title style={font=\small\bfseries}, axis lines*=left, tick align=outside
]
\addplot[gray,dashed,thick,no marks] coordinates {(0.5,7.10) (3.5,7.10)};
\addplot[blue!65!black,only marks,mark=*,mark size=2.6pt]
  coordinates {(0.89,14.03) (1,13.61) (1.11,13.00)};
\addplot[orange!75!black,only marks,mark=square*,mark size=2.4pt]
  coordinates {(1.92,3.07) (2.08,3.40) (2.92,3.71) (3.08,3.65)};
\end{axis}
\end{tikzpicture}
\end{minipage}\hfill
\begin{minipage}[t]{0.47\linewidth}
\centering
\begin{tikzpicture}
\begin{axis}[
  width=\linewidth, height=4.9cm, title={Math (50 updates)},
  ylabel={Mean accuracy (\%)}, xmin=0.5, xmax=2.5, ymin=40, ymax=60,
  xtick={1,2}, xticklabels={Future self,32B}, ytick={40,45,50,55,60},
  tick label style={font=\small}, label style={font=\small},
  title style={font=\small\bfseries}, axis lines*=left, tick align=outside
]
\addplot[gray,dashed,thick,no marks] coordinates {(0.5,47.4) (2.5,47.4)};
\addplot[blue!65!black,only marks,mark=*,mark size=2.6pt]
  coordinates {(0.94,54.9) (1.06,55.6)};
\addplot[orange!75!black,only marks,mark=square*,mark size=2.4pt]
  coordinates {(1.94,47.9) (2.06,48.6)};
\end{axis}
\end{tikzpicture}
\end{minipage}
\caption{Student scores after swapping the OPD teacher. Each point is a training seed. The dashed lines mark the starting students (7.10 on code, 47.4\% on math). Circles use the student's own RL-trained descendant. Squares use teachers with higher standalone scores. Code averages five attempts on each of 172 problems. Math averages 16 attempts on each of 143 problems. The score scales are specific to each benchmark.}
\label{fig:matched}
\end{figure}

Our code benchmark comprises 172 held-out FrontierCS algorithmic problems \citep{frontiercs2025}. Each attempt is one-shot: the model reasons for up to 98,304 tokens and writes a complete C++ program without tools or execution feedback. The grader awards partial credit. We evaluate five attempts per problem and distinguish mean score ($\avgk{5}$) from best-of-five score ($\bestk{5}$) and coverage. We also repeat the teacher swap with publicly available math models (\cref{fig:matched}).

\paragraph{Findings.}
\begin{enumerate}[leftmargin=*,itemsep=2pt]
  \item \textbf{The choice of teacher reversed the code result.} After ten updates, the 9B student rose from 7.10 to 13.55 under its future self (three-seed mean) and fell below 4 under either stronger teacher (\cref{sec:matched}). At similar student training budgets, direct RL scored 10.17 (41.7M tokens, estimated) and OPD scored 14.03 (35.3M, seed~0). Training the future-self teacher first took about 747M additional tokens (\cref{sec:budget}).
  \item \textbf{The teacher ordering replicated on math.} After 50 OPD updates from the same 1.5B student, the future-self arm scored 55.2\% and the 32B-teacher arm 48.2\%, against a 47.4\% starting accuracy (two seeds per arm, \cref{sec:replication}). All matched-swap results are evaluated directly after OPD, without subsequent RL.
  \item \textbf{Adding rewards from test cases gave no clear gain with Flash.} The student scored 6.88 after learning from Flash's example solutions. Five updates with Flash scoring the student's own attempts added 0.84 points on average across three seeds. We continued from the highest-scoring seed (8.86) and added rewards for how well its programs ran. It reached 9.21 and finished at 8.83. A run that alternated learning from examples, Flash feedback and test scores ended at 9.18 (\cref{sec:mixed}).
  \item \textbf{The better teacher found the student's reasoning less surprising.} In both swaps, the student learned more from the teacher that assigned higher probability, on average, to the reasoning tokens the student wrote before training. Both winning teachers were later versions of their own students. The math teacher had also practiced on the problems used for distillation. We have yet to test whether a separately trained teacher that scores the student's reasoning just as highly can teach it as well.
\end{enumerate}

Predicting how long the student would reason proved harder. At points where the student had already drafted its program, we measured how likely each teacher was to stop reasoning there. We used those probabilities to predict whether training would make the student's reasoning longer or shorter. Only one of four predictions was right. In a separate run with the 27B teacher, the student still used a median of 51{,}000 reasoning tokens per attempt, yet its programs scored worse than before training (\cref{sec:lengthterm,sec:prereg}).

\section{Related Work}
\label{sec:related}

\paragraph{When a stronger solver makes a weaker teacher.} This distinction predates language models. In image classification, stopping a teacher's training early could improve the accuracy of the student it taught \citep{cho2019efficacy}. In math OPD, \citet{rethinkingopd2026} found that a 1.5B student learned from its own RL-trained descendant and gained little from a stronger 7B teacher. They also compared two separately trained 4B teachers for a 1.7B student. The teacher that initially shared more of the student's likely next tokens produced the better student. We test teacher choice on long, one-shot programming problems with partial credit and make a prospective prediction on math using DeepScaleR and a stronger 32B teacher.

\paragraph{Choosing a teacher before training.} \citet{yang2026entropyshape} compared where a student and a teacher were uncertain about the next token. This measure identified the best of five teachers for each of two students. A different comparison shows a limit to using agreement alone. For two students studied by \citet{synergy2026}, the 32B teacher had the smallest initial reverse KL divergence, yet another teacher produced higher accuracy. Our ranking asks how much probability each candidate assigns to the reasoning the starting student actually wrote. We test which teacher it selects and whether that teacher's advantage carries over to another student.

\paragraph{What the teacher sees.} The Qwen3 team trained smaller models first on larger models' answers and then on the students' own responses with feedback from those teachers \citep{qwen3report2025}. Our Flash experiments test this sequence starting from a released, post-trained student (\cref{sec:flash}). A separate issue arises when models use different tokenizers. Teacher probabilities can be mapped onto student tokens through their shared byte prefixes \citep{wang2026byte}. Our matched comparisons use shared token IDs, so each teacher can score the same student choice directly.

\paragraph{When reasoning ends.} \citet{eosdisagree2026} traced excessive response length to teachers and students assigning stopping probability to different end-of-sequence tokens. Our code models share the token that ends the reasoning block. We test whether the probability they assign to that token predicts when the student will stop after training (\cref{sec:stopping,sec:prereg}).

\section{Experimental Setup}
\label{sec:setup}

\subsection{Task and evaluation}
\label{sec:eval}

We evaluated the student on 172 held-out FrontierCS programming problems. Each attempt starts with one problem statement. The model can use up to 98,304 tokens to reason and 32,768 to write one complete C++ program. If it reaches the reasoning limit, the harness ends the thinking block and asks for the program. We call this a \emph{forced close}. The grader runs the submitted program on the full test set and awards partial credit on a 0--100 scale.

Our primary score, $\avgk{5}$, is the mean of five attempts per problem, averaged over problems. We also report $\bestk{5}$, the mean of each problem's highest score, and \emph{coverage}, the number of problems earning any credit across five attempts. Compilation and forced-close rates describe what happened on individual attempts. We evaluate student checkpoints with the same prompt, sampler and grader. Unless otherwise stated, 95\% intervals are paired by problem (\cref{app:code-eval}).

\subsection{Models}
\label{sec:models}

\begin{table}[htbp]
\centering
\small
\caption{Models and checkpoints. The first five rows link to the released models. The last three are checkpoints we trained. Scores are $\avgk{5}$ / $\bestk{5}$ on the 172 tasks.}
\label{tab:models}
\setlength{\tabcolsep}{4pt}
\begin{tabular}{@{}llll@{}}
\toprule
Symbol & Model & Role & Score \\
\midrule
$\piz$ & \href{https://huggingface.co/Qwen/Qwen3.5-9B}{Qwen3.5-9B} (post-trained release) & student; teacher's RL init. & 7.10 / 15.12 \\
$\pibase$ & \href{https://huggingface.co/Qwen/Qwen3.5-9B-Base}{Qwen3.5-9B-Base} (pretrained) & SFT-from-base study (\cref{sec:warmup}) & 7.99 / 17.57 \\
$\Tf$ & \href{https://huggingface.co/Qwen/Qwen3.8-Flash-Next-FP8}{Qwen3.8-Flash-Next-FP8} (125B MoE, ${\sim}$6B active) & strong distant teacher & 34.99 / 61.40 \\
--- & \href{https://huggingface.co/Qwen/Qwen3.8-27B}{Qwen3.8-27B} & distant teacher; demonstrations & 30.42 / 57.36 \\
--- & \href{https://huggingface.co/moonshotai/Kimi-K3}{Kimi K3} & demonstrations (\cref{sec:nll}) & --- \\
\midrule
$\Trl$ & $\piz$ + 124 outcome-RL updates & future-self teacher & 14.65 / 21.65 \\
$\pisftf$ & $\piz$ + 3-epoch SFT on $\Tf$ traces & student for Flash OPD & 6.88 / 20.20 \\
$\pisftk$ & $\piz$ + 1-epoch SFT on K3 traces & SFT comparison & 0.33 / 1.29 \\
\bottomrule
\end{tabular}
\end{table}

The released 9B model $\piz$ is the starting student (\cref{tab:models}). Its future-self teacher $\Trl$ is the same model after 124 updates of outcome RL. Flash and the 27B teacher each scored higher on their own code evaluations than the future self (34.99 and 30.42 vs.\ 14.65) and earned credit on more problems (152 and 143 vs.\ 105). The distant teachers used their own evaluation settings. The matched comparison is between the students they trained (\cref{sec:matched,app:code-eval}).

\subsection{On-policy distillation objective}
\label{sec:objective}

Let the student $\pi_\theta$ sample a response $x_{1:L}$ to prompt $q$. A frozen teacher $T$ scores every sampled token under the same prefix. The per-token signal is
\begin{equation}
  d_t \;=\; \log p^{(\tau)}_T(x_t \mid q, x_{<t}) \;-\; \log \pi_\theta(x_t \mid q, x_{<t}),
  \label{eq:dt}
\end{equation}
where $p^{(\tau)}_T \propto p_T^{1/\tau}$ is the teacher distribution at scoring temperature $\tau$. A positive $d_t$ means the teacher put more probability than the student on the token the student just chose. At a fixed prefix, the expected value of $d_t$ under the student is the negative reverse KL divergence from the student to the teacher \citep{lu2025opd}.

For the code teacher ranking, we scored the same starting student's reasoning tokens with each teacher at its native temperature ($\tau=1$). Subtracting the student's log probability shifts every teacher's score by the same amount. Ranking by mean teacher log probability therefore gives the same order as ranking by the mean signed gap $d_t$ (\cref{sec:proximity}). During the matched code training runs we used $\tau=0.8$.

To give an earlier choice some credit for what follows it, we optionally add a discounted sum of later teacher signals, the \emph{return},
\begin{equation}
  A_t \;=\; d_t \;+\; \lambda \sum_{k \ge 1} \gamma^{k} d_{t+k},
  \label{eq:return}
\end{equation}
and train on $A_t$ with a policy-gradient update. A reference penalty $\beta\,\mathrm{KL}(\pi_\theta\|\pi_{\mathrm{ref}})$ limits drift from the starting student. The teacher signal covers reasoning and answer tokens. We mask special targets, including the token that ends reasoning, so its probability can be studied as a diagnostic (\cref{sec:stopping}). The loss calculation and trainer checks are in \cref{app:loss,sec:checks}.

\paragraph{Where execution rewards enter.} In the matched swaps the teacher supplies the entire task signal, and we evaluate the student immediately after OPD. The optional return in \cref{eq:return} sums later teacher log-probability signals on the same completion. A program's execution reward comes from running it through the grader. We tested adding that reward to the Flash training objective and, separately, alternating OPD with outcome-RL stages (\cref{sec:mixed}).

\subsection{Training the future-self teacher}
\label{sec:teacher}

We trained a copy of the released 9B student with GRPO \citep{shao2024deepseekmath}, rewarding its programs for their execution scores. The first 40 updates used synthetic competitive programming problems \citep{frontiersmith2026}. The final 84 used a mix of synthetic and AtCoder Heuristic Contest problems. An LLM judge also helped rank programs during 69 of the 124 updates. Screening the training problems against FrontierCS found no matches (\cref{app:training}).

Training this teacher cost about 747M trainer tokens and an estimated 1,243 GPU-hours. For OPD we returned to the original 9B student and used the trained copy only to score its responses. We account separately for building the teacher and distilling from it (\cref{sec:budget}).

\section{Matched Teacher Swaps on Code}
\label{sec:matched}

Every run in the code panel of \cref{fig:matched} starts from the released 9B student $\piz$. We changed which frozen teacher scored the tokens that student generated. The candidates were $\piz$'s own checkpoint after 124 outcome-RL updates ($\Trl$), Flash ($\Tf$), and Qwen3.8-27B. All three share the student's tokenizer. In seed~0 we used the same task draw at every update across all three teachers. Each student generated fresh responses as its weights changed. We ran three seeds with $\Trl$ and two with each distant teacher. We evaluated all seven checkpoints immediately after ten OPD updates, with five attempts on each of the same 172 held-out tasks.

Each update used the same adapter architecture (rank-64 LoRA \citep{hu2022lora}), learning rate and teacher temperature across the three teachers. All three arms used the teacher return and zero execution reward. The full settings are in \cref{sec:compute}.

\paragraph{Student scores after the swap.} The future self raised the student's mean score from 7.10 to 13.55 across three seeds (range 13.00--14.03). With Flash, the two students scored 3.07 and 3.40. With the 27B, they scored 3.71 and 3.65. All four losses relative to the starting student had paired intervals below zero. With either distant teacher the student also earned credit on fewer problems (47--54, down from 70) and its $\bestk{5}$ fell from 15.12 to 8.0--9.1 (\cref{tab:main}). The paired comparisons across seeds and teachers are in \cref{app:code-contrasts}.

\begin{table}[htbp]
\centering
\footnotesize
\setlength{\tabcolsep}{4pt}
\caption{Held-out code scores on 172 tasks with five attempts per task (0--100 scale). The middle block shows the matched swaps after ten updates from $\piz$. Mean score under the future self was \textbf{13.55} across three seeds (range 13.00--14.03). The bottom block gives the separate five-update, full-weight Flash experiment from $\pisftf$ (\cref{sec:flash}). Each difference is paired against that row's starting student. $^{*}$ marks 95\% intervals excluding zero. Median thinking is measured on the held-out attempts. Dashes mark entries without a recorded value or an applicable comparison.}
\label{tab:main}
\begin{tabular}{@{}llccccc@{}}
\toprule
Policy & Teacher & $\avgk{5}$ & $\bestk{5}$ & Coverage & Median thinking & $\Delta\avgk{5}$ \\
\midrule
$\piz$ (released) & --- & 7.10 & 15.12 & 70 & 56.9K & --- \\
$\Trl$ ($\piz$ + 124 RL updates) & --- & 14.65 & 21.65 & 105 & 61.7K & --- \\
\midrule
OPD from $\piz$, seed 0 & $\Trl$ & 14.03 & 22.71 & 105 & 54.2K & $+6.94^{*}$ \\
\quad seed 1 & $\Trl$ & 13.61 & 22.18 & 108 & 52.9K & $+6.51^{*}$ \\
\quad seed 2 & $\Trl$ & 13.00 & 21.91 & 101 & 50.6K & $+5.90^{*}$ \\
OPD from $\piz$, seeds 0 / 1 & $\Tf$ & 3.07 / 3.40 & 8.72 / 8.03 & 49 / 47 & 1.4K / 1.6K & $-4.03^{*}$ / $-3.69^{*}$ \\
OPD from $\piz$, seeds 0 / 1 & Qwen3.8-27B & 3.71 / 3.65 & 8.91 / 9.14 & 50 / 54 & 3.0K / 3.8K & $-3.39^{*}$ / $-3.45^{*}$ \\
\midrule
$\pisftf$ ($\piz$ + Flash SFT) & --- & 6.88 & 20.20 & 86 & --- & --- \\
OPD from $\pisftf$, seed 0 & $\Tf$ & 8.86 & 20.82 & 86 & --- & $+1.98^{*}$ \\
\quad seeds 1 / 2 & $\Tf$ & 6.84 / 7.47 & 17.05 / 17.12 & 83 / 81 & 98.3K / 98.3K & $-0.05$ / $+0.59$ \\
\bottomrule
\end{tabular}
\end{table}

\paragraph{Earlier checkpoints.} By update two, both distant-teacher students in seed~0 already scored below their starting value of 7.10. Flash scored 3.85 (paired $\Delta=-3.25$, 95\% CI $[-5.21,-1.50]$) and the 27B scored 4.23 ($-2.87$, $[-4.38,-1.52]$). At update four the 27B student scored 2.67 ($-4.43$, $[-6.08,-2.96]$), 1.04 points below its own update-ten score (CI $[-2.01,-0.23]$ for update four minus ten).

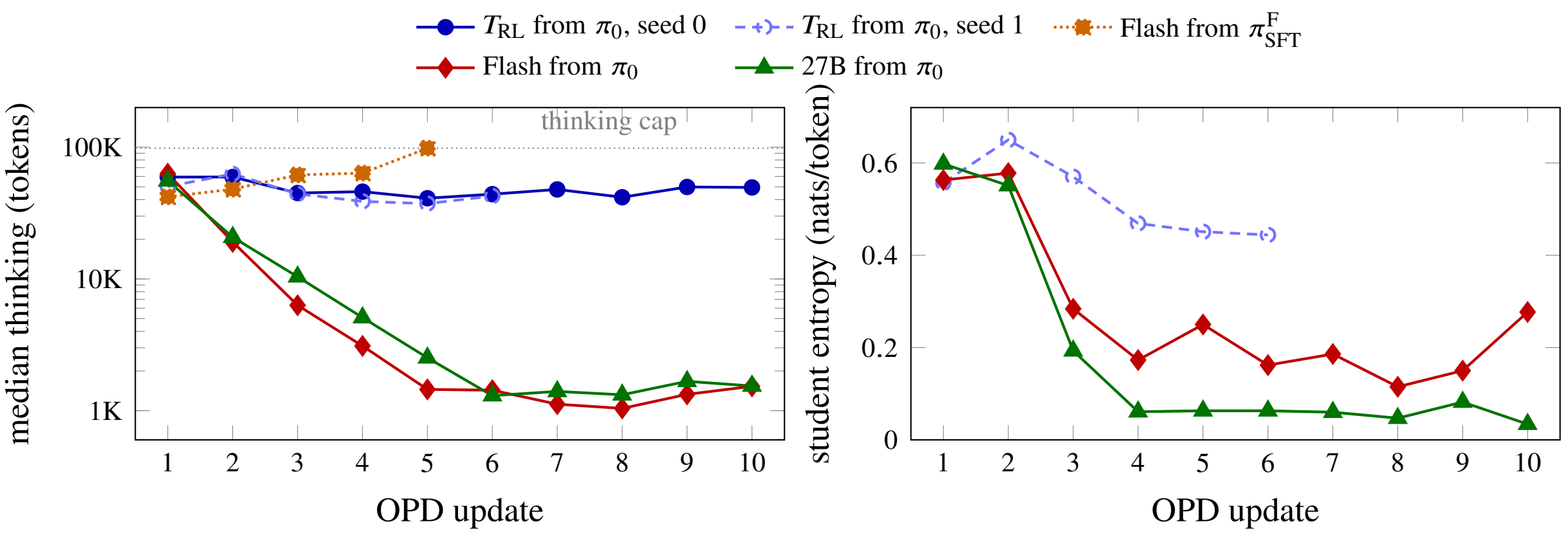
\begin{figure}[t]
\centering
\begin{tikzpicture}
\begin{axis}[
  name=len, width=0.5\linewidth, height=5cm,
  xlabel={\small OPD update}, ylabel={\small median thinking (tokens)},
  xmin=0.5, xmax=10.5, xtick={1,...,10},
  ymode=log, ymin=0.6, ymax=200, ytick={1,10,100}, yticklabels={1K,10K,100K},
  tick label style={font=\scriptsize}, label style={font=\small},
  legend columns=3, legend cell align=left,
  legend style={at={(1.12,1.04)},anchor=south,font=\scriptsize,draw=none,/tikz/every even column/.append style={column sep=6pt}},
]
\addplot[mark=*,blue!70!black,thick] coordinates {(1,59.3) (2,59.4) (3,44.9) (4,46.0) (5,41.0) (6,44.0) (7,47.8) (8,41.7) (9,49.9) (10,49.5)};
\addlegendentry{$T_{\mathrm{RL}}$ from $\pi_0$, seed 0}
\addplot[mark=o,blue!55,thick,densely dashed] coordinates {(1,50.3) (2,62.7) (3,44.2) (4,38.8) (5,37.3) (6,42.6)};
\addlegendentry{$T_{\mathrm{RL}}$ from $\pi_0$, seed 1}
\addplot[mark=square*,orange!80!black,thick,densely dotted] coordinates {(1,42.0) (2,48.0) (3,61.6) (4,63.6) (5,98.3)};
\addlegendentry{Flash from $\pi^{\mathrm{F}}_{\mathrm{SFT}}$}
\addplot[mark=diamond*,mark size=2.6pt,red!75!black,thick] coordinates {(1,62.96) (2,19.10) (3,6.31) (4,3.10) (5,1.45) (6,1.43) (7,1.12) (8,1.04) (9,1.33) (10,1.53)};
\addlegendentry{Flash from $\pi_0$}
\addplot[mark=triangle*,mark size=2.6pt,green!45!black,thick] coordinates {(1,55.56) (2,20.76) (3,10.39) (4,5.09) (5,2.52) (6,1.30) (7,1.40) (8,1.32) (9,1.67) (10,1.54)};
\addlegendentry{27B from $\pi_0$}
\addplot[gray,densely dotted,domain=0.5:10.5,forget plot] {98.3};
\node[font=\scriptsize,gray,anchor=south] at (axis cs:7.8,98.3) {thinking cap};
\end{axis}
\begin{axis}[
  at={(len.east)}, anchor=west, xshift=1.3cm, width=0.5\linewidth, height=5cm,
  xlabel={\small OPD update}, ylabel={\small student entropy (nats/token)},
  xmin=0.5, xmax=10.5, ymin=0, ymax=0.72, xtick={1,...,10},
  tick label style={font=\scriptsize}, label style={font=\small},
]
\addplot[mark=o,blue!55,thick,densely dashed] coordinates {(1,0.557) (2,0.650) (3,0.570) (4,0.469) (5,0.451) (6,0.444)};
\addplot[mark=diamond*,mark size=2.6pt,red!75!black,thick] coordinates {(1,0.563) (2,0.578) (3,0.284) (4,0.173) (5,0.250) (6,0.162) (7,0.186) (8,0.115) (9,0.150) (10,0.277)};
\addplot[mark=triangle*,mark size=2.6pt,green!45!black,thick] coordinates {(1,0.598) (2,0.551) (3,0.193) (4,0.061) (5,0.063) (6,0.063) (7,0.060) (8,0.047) (9,0.082) (10,0.034)};
\end{axis}
\end{tikzpicture}
\caption{Training-time statistics per OPD update (8 tasks $\times$ 8 samples; log scale on the left). From the released student $\pi_0$ with identical recipe and task draws, the future-self teacher keeps reasoning stable, while both distant teachers collapse it to about 1.5K tokens within five or six updates and drive student entropy down. Flash on its warmed-up student $\pi^{\mathrm{F}}_{\mathrm{SFT}}$ (full-weight recipe, \cref{sec:flash}) fails the other way, running to the 98K cap. Entropy is not available per update for the Flash-route run and $T_{\mathrm{RL}}$ seed 0.}
\label{fig:dynamics}
\end{figure}

\paragraph{Reasoning length.} Under the distant teachers, median thinking during training fell from 56--63K tokens on the first update to about 1.5K by update five or six and stayed there (\cref{fig:dynamics}). On the held-out tasks, seed~0 used a median of 1.4K tokens with Flash and 3.0K with the 27B. Seed~1 used 1.6K and 3.8K, respectively.

Under the distant teachers, student entropy fell from about 0.6 to 0.03--0.28. From update four onward the 27B-taught student had lower entropy than its teacher. By the final update the fraction of tokens on repeated lines was 31.7\% in seed~0 and 56.7\% in seed~1. The short rollouts typically contained a few thousand tokens of analysis followed by code with baseline-level performance. Under $\Trl$, median thinking held between 37K and 63K tokens in both seeds with per-update logs, and seed~1 entropy stayed at 0.44--0.65.

Both distant teachers assigned higher closing probabilities than the student at points where its reasoning contained code closely matching its final, compiling program (\cref{sec:stopping}). In a subsequent pre-registered test, that probe predicted the direction of length change in one of four code runs (\cref{sec:prereg}).

\subsection{Transfer from the RL-trained future self}
\label{sec:futureself}

The students recovered an average of 85\% of their teacher's $+7.55$-point gain over $\piz$ (\cref{tab:main}). Across three seeds the mean gain was $+6.45$ on $\avgk{5}$ (95\% CI $[+4.58,+8.54]$ resampling tasks and seeds) and $+7.15$ on $\bestk{5}$ ($[+4.14,+10.28]$). The students finished an average of 1.10 points below the teacher (CI $[-2.42,+0.11]$). Seed~0 earned credit on 105 tasks, up from 70 and equal to the teacher's coverage, although its programs compiled less often (80.1\% vs.\ 95.0\%). The temperature controls in \cref{sec:entropy} test how much of the gain sharpening alone can explain.

\paragraph{Reasoning length during training.} Under the future self, forced closes fell from 23\% to 0--8\% after the first update. Median reasoning stayed between 37K and 63K tokens in the two seeds with per-update logs. Training-time compilation rose from 47\% to 88\%, and mean $|d_t|$ at $\tau=0.8$ fell from 0.136 to 0.062 nats per token while the KL to $\piz$ grew to 0.033. The full-weight Flash run from $\pisftf$ reached the 98K cap by update five (\cref{fig:dynamics}). On the same problems it used 36,654 more thinking tokens on average than $\pisftf$ (95\% CI $[25{,}020,47{,}983]$), with 53\% forced closes against 6\% for its start.

\paragraph{Agreement after training.} Across tasks, the partial correlation between the seed-0 student's and $\Trl$'s mean scores, controlling for $\piz$, was 0.78 (95\% CI $[0.63,0.88]$). On prefixes generated by the trained student, its estimated KL to the teacher was 0.0194 nats per token, compared with 0.0245 for the starting student on those same prefixes. On the original student's prefixes the estimates were closer (0.0193 vs.\ 0.0200). These are native-temperature measurements. \Cref{app:diagnostics} gives the estimator, uncertainty and reverse-direction comparisons.

In weight space, 0.39\% of the LoRA update's norm lay in the top-64 singular subspace of the RL weight change $\Trl-\piz$, about 19 times the expected share for a random subspace of that rank (about 0.02\%).

\subsection{What it took to reach the teacher}
\label{sec:budget}

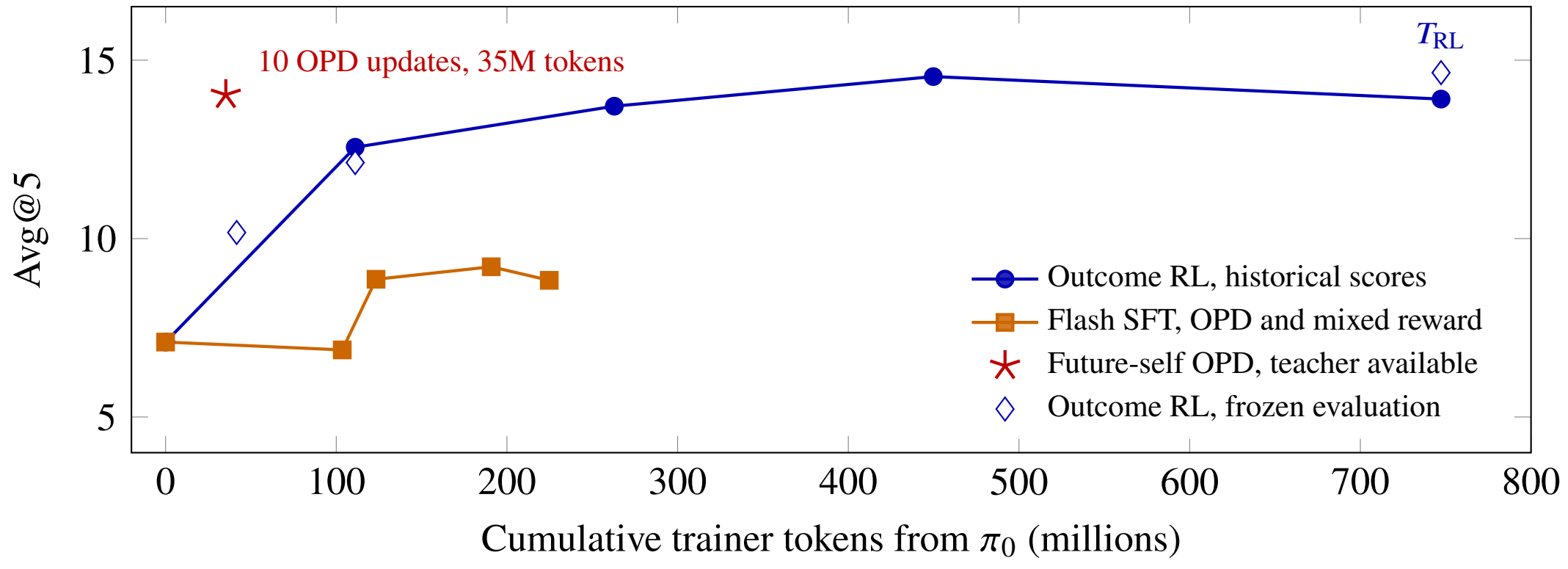
\begin{figure}[t]
\centering
\begin{tikzpicture}
\begin{axis}[
  width=0.86\linewidth, height=5.6cm,
  xlabel={Cumulative trainer tokens from $\pi_0$ (millions)}, ylabel={$\mathrm{Avg@}5$},
  xmin=-20, xmax=800, ymin=4, ymax=16.5,
  legend cell align=left,
  legend style={at={(0.98,0.03)},anchor=south east,font=\scriptsize,draw=none,fill=white,fill opacity=0.85,text opacity=1},
  tick label style={font=\small}, label style={font=\small},
]
\addplot[mark=*,color=blue!70!black,thick] coordinates {(0,7.10) (111.1,12.56) (262.9,13.71) (449.8,14.54) (747.3,13.91)};
\addlegendentry{Outcome RL, historical scores}
\addplot[mark=square*,color=orange!80!black,thick] coordinates {(0,7.10) (103.5,6.88) (123.3,8.86) (190.8,9.21) (224.8,8.83)};
\addlegendentry{Flash SFT, OPD and mixed reward}
\addplot[only marks,mark=star,mark size=4pt,color=red!75!black,thick] coordinates {(35.3,14.03)};
\addlegendentry{Future-self OPD, teacher available}
\addplot[only marks,mark=diamond*,mark size=3pt,mark options={draw=blue!70!black,fill=white},color=blue!70!black] coordinates {(41.6544,10.17) (111.1,12.13) (747.3,14.65)};
\addlegendentry{Outcome RL, frozen evaluation}
\node[font=\scriptsize,anchor=west,color=red!75!black] at (axis cs:45,14.9) {10 OPD updates, 35M tokens};
\node[font=\scriptsize,anchor=south,color=blue!70!black] at (axis cs:747.3,14.9) {$T_{\mathrm{RL}}$};
\end{axis}
\end{tikzpicture}
\caption{Code score against cumulative student trainer tokens from $\pi_0$. Open diamonds are frozen evaluations of outcome RL after 15, 40 and 124 updates (10.17, 12.13 and 14.65). The blue line shows earlier evaluations along the same trajectory. Token counts for the first 40 RL updates are estimated. The star marks future-self OPD from an existing teacher (seed~0). The orange line follows Flash SFT, then OPD and outcome-dominated hybrid training (\cref{sec:flash}). Its 8.86 point is seed~0.}
\label{fig:budget}
\end{figure}

\begin{table}[t]
\centering
\small
\caption{Training budgets. Tokens are trainer tokens. GPU-hours are wall-clock time $\times$ allocated GPUs, including teacher servers. ``est.'' marks estimates.}
\label{tab:budget}
\begin{tabular}{@{}lccc@{}}
\toprule
Route & Updates & Tokens & GPU-hours \\
\midrule
Outcome RL, $\piz \to \Trl$ & 124 & 747M (111M est.) & ${\approx}$1,243 (est.) \\
Future-self OPD, $\piz \to$ student & 10 & 35.3M & ${\approx}$56.6 (est.) \\
\quad teacher scoring (prefill) & --- & 36.6M & incl.\ above \\
\midrule
Flash SFT, $\piz \to \pisftf$ & 207 & 103.5M & ${\approx}$15 \\
Flash OPD, $\pisftf \to$ student & 5 & 19.9M & ${\approx}$50 \\
\quad teacher scoring (prefill) & --- & ${\approx}$20.4M (est.) & incl.\ above \\
\bottomrule
\end{tabular}
\end{table}

Creating the teacher took about 747M trainer tokens and an estimated 1,243 GPU-hours. Once the teacher existed, the seed-0 student used 35.3M trainer tokens over ten updates, plus 36.6M teacher-prefill tokens (\cref{tab:budget}). The allocation was about 57 GPU-hours.

For a direct comparison, we selected the saved outcome-RL checkpoint nearest the OPD token budget before reading its score. Per-update token logs for the first stage were unavailable, so we estimated cumulative counts from the stage mean. Update~15 was nearest at 41.65M tokens, 17.8\% above the OPD budget. It scored 10.17 on the frozen evaluation, gaining $+3.07$ over $\piz$ (95\% CI $[+1.59,+4.63]$). The 35.35M-token OPD student scored 14.03, a paired advantage of $+3.86$ ($[+2.07,+5.77]$). The $\bestk{5}$ difference was $+1.95$ ($[-0.72,+4.71]$). This comparison measures the cost of transferring an existing teacher's performance. Training the teacher and scoring its prefixes are separate costs.

\section{Adapting Distillation to a Stronger, More Distant Teacher}
\label{sec:distant}

Flash and the 27B model scored 34.99 and 30.42 on FrontierCS, respectively, yet both reduced the released student's score in the matched swap (\cref{sec:matched}). We tested whether demonstrations could first bring the student closer to the teacher, whether program execution rewards could supplement the token signal, and whether shorter teacher reasoning or explicit hints helped. These experiments varied the recipe and budget. Each comparison below identifies the student and evaluation it used.

\subsection{Flash on a warmed-up student}
\label{sec:flash}

We first trained the released 9B student on 1,110 Flash demonstrations and then used Flash to score that student's own attempts. Each comparison in \cref{tab:flash} starts from this same student, $\pisftf$, and updates all its weights for five steps. We varied the teacher temperature and whether later tokens contributed to the return. The upper panel reports one attempt per task and the lower panel reports five. Training settings are in \cref{sec:compute}.

\begin{table}[t]
\centering
\small
\caption{Flash OPD from $\pisftf$. One-sample rows (top) are separate screens on the 172 tasks; five-sample rows (bottom) are the paired main evaluation. ``Return'' adds the return of \cref{eq:return} with $\gamma = 0.9999$.}
\label{tab:flash}
\begin{tabular}{@{}lcccc@{}}
\toprule
Recipe & Attempts/task & Mean score & $\bestk{5}$ & Forced close \\
\midrule
$\pisftf$ (start) & 1 & 7.58 & --- & 26.7\% \\
Teacher at $\tau{=}1.0$ & 1 & 3.64 & --- & 45.9\% \\
Teacher at $\tau{=}1.0$ + outcome reward & 1 & 4.43 & --- & 58.7\% \\
Teacher at $\tau{=}0.8$ & 1 & 7.31 & --- & 73.8\% \\
Teacher at $\tau{=}0.8$ + return ($\lambda{=}0.001$) & 1 & 6.78 & --- & 65.7\% \\
\midrule
$\pisftf$ (start) & 5 & 6.88 & 20.20 & 23.6\% \\
Teacher at $\tau{=}0.8$ & 5 & 7.09 & 15.47 & ${\sim}$72\% \\
Teacher at $\tau{=}0.8$ + return ($\lambda{=}0.01$), seed 0 & 5 & 8.86 & 20.82 & 57.4\% \\
\quad same recipe, seeds 1 / 2 & 5 & 6.84 / 7.47 & 17.05 / 17.12 & 56.4 / 54.4\% \\
\bottomrule
\end{tabular}
\end{table}

\paragraph{Length drift after entropy matching.} Scoring with $\Tf$ at $\tau = 1$ cut the one-sample score from 7.58 to 3.64 and lengthened median thinking from 57.5K to 94.3K tokens. On student prefixes, raw $\Tf$ had about 0.11 nats more entropy per token than the student, and $\tau = 0.8$ matched them (0.415 vs.\ 0.428 nats). Thinking still grew. By update five training entropy had collapsed to about 0.11, $\bestk{5}$ had dropped by 4.72 (CI $[-8.54, -1.06]$) and about 72\% of attempts hit the cap.

\paragraph{What the return changed.} With the return at $\lambda=0.01$, the three seeds reached 8.86, 6.84 and 7.47 from a starting $\avgk{5}$ of 6.88. The mean gain was $+0.84$ (95\% CI $[-0.76,+2.54]$ resampling tasks and seeds). Only the first seed showed a clear gain ($+1.98$, CI $[+0.51,+3.62]$). Its coverage stayed the same and $\bestk{5}$ rose by 0.62. Across seeds, the mean change in $\bestk{5}$ was $-1.87$ (CI $[-5.52,+1.90]$). Task draws and the teacher serving configuration both varied across seeds (\cref{app:code-contrasts,sec:checks}).

The return delayed the rise in reasoning length. Median thinking at the second training batch was about 48K tokens with the return and 84K without it, but reached the cap by the fifth batch. On evaluation, all three seeds forced a close on 54--57\% of attempts, up from 23.6\%. In seed~0, mean score rose from 8.25 to 15.96 on attempts that ended reasoning naturally and from 2.44 to 3.60 on forced closes. About 62\% of the gain came from tasks on which the starting student already reasoned for fewer than 16K tokens.

\paragraph{Changing which tokens Flash supervises.} We trained about 30 Flash-route configurations. Across these runs, reasoning grew toward the cap whenever Flash scored thinking tokens for at least two updates without a dominant competing objective. With forward KL on Flash's top 20 tokens, median reasoning grew by 23.0K tokens in five updates. Masking choices, length rules and a stronger reference penalty also failed to control length in the tested settings (\cref{app:flash-controls}).

Restricting per-token Flash feedback to the student's final answer raised compilation from 69.1\% to 73.6\% at update four (paired gain 4.5 percentage points, CI $[0.3,8.8]$). Mean score reached 7.86 from 6.88 ($+0.98$, CI $[-0.34,+2.32]$), and $\bestk{5}$ was 19.48 against 20.20. Flash supplied only 0.2--0.9\% of the measured update in this arm.

We also reduced Flash feedback to one score per completion. After four updates, the whole-rollout version (teacher weight 2) scored 7.05 in the one-attempt screen, against 7.58 for the warmed student ($-0.54$, CI $[-4.07,+2.98]$). Both forced a close on 26.7\% of attempts. Computing a single score from answer tokens alone (weight 4) gave 6.09 ($-1.49$, CI $[-4.48,+1.35]$) and lowered compilation by 9.9 percentage points (CI for the difference $[-19.2,-0.6]$). Adding a supervised likelihood term on teacher-written answers yielded one-attempt scores of 8.16, 5.78 and 5.24 at updates 10, 20 and 30.

\subsection{Mixing Flash feedback with program execution rewards}
\label{sec:mixed}

The grader can supply partial credit for a working program even when Flash assigns low probability to the reasoning that produced it. In the five-update screen at teacher temperature 1, the warmed student scored 7.58 before training, 3.64 with Flash alone and 4.43 with Flash plus the execution reward (one attempt per problem, \cref{tab:flash}). The difference was 0.79 points in favor of mixing, with both scores below the starting checkpoint.

A separate pilot applied Flash feedback to early reasoning and code with fixed weights of 1.0 on Flash and 0.34 on the outcome term. By update six it forced a close on 65.6\% of its training attempts. No held-out score was recorded for that pilot.

\begin{table}[htbp]
\centering
\small
\caption{Adding program rewards after Flash SFT and OPD. Each checkpoint was evaluated on the same 172 problems with five attempts per problem. Tokens are cumulative student trainer tokens from $\piz$ and exclude teacher prefill and demonstration generation. Coverage counts problems receiving positive credit in at least one attempt. Paired differences in the text use unrounded scores.}
\label{tab:mixed}
\begin{tabular}{@{}lrrrr@{}}
\toprule
Checkpoint on the Flash route & Trainer tokens & $\avgk{5}$ & $\bestk{5}$ & Coverage \\
\midrule
After Flash SFT & 103.5M & 6.88 & 20.20 & 86 \\
After OPD, update 5 (seed 0) & 123.3M & 8.86 & 20.82 & 86 \\
Mixed objective, update 20 & 190.8M & 9.21 & 19.39 & 91 \\
Mixed objective, update 30 & 224.8M & 8.83 & 17.61 & 86 \\
\bottomrule
\end{tabular}
\end{table}

In the longer continuation, the configured Flash weight decreased from 0.5 toward 0.3 over 20 updates starting at update six. The execution term used the complementary weight multiplied by 6.33. This continuation used the highest-scoring of the three update-five OPD checkpoints. It beat the SFT-only student at updates 20 and 30 ($+2.33$, CI $[+0.54,+4.32]$, and $+1.95$, CI $[+0.31,+3.73]$). Relative to the 8.86 checkpoint where mixing began, the changes were $+0.35$ ($[-1.18,+1.99]$) and $-0.02$ ($[-1.39,+1.35]$). Best-of-five score declined at both checkpoints (\cref{tab:mixed}).

Plain outcome RL from the same Flash-warmed student reached 8.12 after 18 updates and 76.7M additional trainer tokens. At update 20 the hybrid had used 87.4M tokens after SFT and was 1.09 points ahead of that RL checkpoint (CI $[-0.33,+2.59]$). Neither this comparison nor the change from update five establishes an incremental benefit from mixing the two signals.

We also tried alternating SFT, OPD and RL as separate training stages. After a 20-update RL stage the student scored 8.34. Five OPD updates later it scored 8.69 ($+0.35$, CI $[-0.78,+1.51]$), while forced closes rose from 32.3\% to 58.4\% and compilation fell from 86.4\% to 77.2\%. The next RL stage brought the student to 9.18 after 396.0M trainer tokens from $\piz$.

Direct outcome RL scored 13.49 after 480.1M tokens (94 updates), a paired difference of $-4.31$ for the alternating route (95\% CI $[-6.54,-2.29]$). The comparison is between saved checkpoints at different budgets. Both evaluations used the same problems and attempts per problem, with different launchers and judging infrastructure.

\subsection{Lowering the teacher's reasoning effort}
\label{sec:effort}

Both distant teachers expose three effort levels: xhigh (the default), medium and low. At xhigh the 27B thought for a median 82K tokens on 2,128 samples of training problems; in a 64-sample pilot, medium and low cut that to 16K and 20K with no significant change in score. Flash at xhigh closed within 64K tokens on none of 10 probe samples; at medium its median was 27.4K, with 83\% natural closes. All Flash demonstrations and OPD runs therefore use medium effort, which also renders exactly the student's prompt tokens. Long demonstrations teach overthinking: full SFT of $\piz$ on 690 xhigh 27B traces (median thinking 89K) gave a half-epoch checkpoint that scored 0.30 $\avgk{1}$, with 17\% compilation.

Lower effort slowed OPD's length drift but did not stop it. We used the 27B as teacher for students fine-tuned on its traces (16 tasks $\times$ 4 samples per update). An xhigh teacher with an xhigh-SFT student sat at the cap from the first update, and compilation reached zero by update 3. With medium effort on both sides (the student fine-tuned on the 459 hinted traces of \cref{sec:nll}, the teacher seeing the same hint), median thinking still doubled in four updates (23.8K to 49.7K), and weighting the thinking-token teacher term by 0.1 did not change that. Only a three times smaller learning rate held it at 21--29K, with a non-significant one-sample gain ($+1.45$, CI $[-1.70, +4.61]$).

\subsection{27B solutions as hints during RL}
\label{sec:hints}

In a 20-update RL pilot starting from $\Trl$, graded 27B solutions supplied a sketch, partial code or a full program on hard training problems, with less help as attempts succeeded. The student wrote its own answer and received the execution score. This pilot had no matched plain-RL control.

With hints removed, it scored 13.02 on a one-sample evaluation against 14.39 for its start (paired $-1.37$, CI $[-3.99, +1.30]$). In 720 unhinted training attempts we found no use of the 27B's core idea, and most hinted successes copied the full program.

\section{Measuring Closeness on the Student's Rollouts}
\label{sec:distance}

Why did the weaker teacher work? We measured each teacher's distance from the student in four ways: overlap with the student's next-token distribution on its own rollouts, entropy on student prefixes, agreement about when to stop reasoning, and, for SFT, how well the student predicts the teacher's text.

\subsection{Gap and overlap}
\label{sec:proximity}

We had each teacher score the same 64 responses sampled from $\piz$ before its first update (\cref{tab:proximity}). The mean teacher--student log-probability gap on sampled tokens is about six times smaller for $\Trl$ than for either distant teacher, and $\Trl$'s top-50 probability mass overlaps $\piz$'s more. Almost every sampled token lies in each teacher's top-50, so the differences come from how the mass is distributed, not from missing support. $\piz$ also predicts $\Trl$'s own thinking far better than the 27B's or $\Tf$'s (0.718 vs.\ 1.291 and 1.708 nats/token), though the distant teachers' answers are more predictable (0.095 and 0.145 vs.\ 0.475).

An earlier checkpoint from the same RL trajectory let us measure proximity at an intermediate teacher strength. After 40 RL updates it scored 12.13 on the frozen evaluation, a gain of $+5.04$ over $\piz$ (95\% CI $[+3.31,+6.92]$). On the same 64 initial rollouts its thinking gap was $-0.0028$ nats per token and its top-50 overlap was 0.9815. The final teacher scored 14.65 with a larger gap ($-0.0208$) and lower overlap (0.9522). The earlier teacher trailed the final teacher by 2.52 points on $\avgk{5}$ (paired 95\% CI $[1.21,3.89]$). A held-out score for the student it taught was unavailable for this comparison.

\paragraph{Mixing the teachers' token probabilities.} On these same rollouts we also scored a geometric mixture, $q_\alpha(x\mid h)\propto p_{\Trl}(x\mid h)^\alpha p_{\Tf}(x\mid h)^{1-\alpha}$, at native temperature and at each student prefix $h$. As the future-self weight rose from 0 through 0.25, 0.50, 0.75 and 1, the mean thinking gaps were $-0.1229$, $-0.0647$, $-0.0327$, $-0.0191$ and $-0.0208$ nats per token. At 0.75 the mixture was slightly closer to the student than the pure future self, including by top-50 overlap (0.9541 vs.\ 0.9522). This was an offline scoring experiment. Neither the mixture's own task performance nor its value as a training teacher was measured.

The disagreement is broad. We split the thinking positions by line type: lines opening with hesitation words such as ``Wait'' or ``Alternatively'', verification, code inside the thinking, repeated lines, positions near the ready point (where the student already has its final program; \cref{sec:stopping}), and the rest. For all three teachers each type's share of the total $|d_t|$ tracks its share of positions, and the distant teachers' mean $d_t$ at $\tau = 0.8$ is two to four times $\Trl$'s in every type. Verification lines carry the most negative signal ($-0.20$ and $-0.16$ for the 27B and $\Tf$, $-0.07$ for $\Trl$) and repeated lines the least. The distant teachers do not object to a particular kind of step; they assign lower probability to the student's text throughout.

Closeness belongs to a teacher--student pair, not to a teacher. On $\pisftf$'s rollouts $\Trl$ is barely closer than the distant teachers (gap $-0.096$ vs.\ $-0.126$ and $-0.129$), about 4.6 times farther than it is from $\piz$.

\begin{table}[t]
\centering
\footnotesize
\caption{Teacher proximity to the student, thinking / code. Gap is the mean of $\log p_T - \log p_S$ on $\piz$'s sampled tokens; it is negative, and we compare gaps by magnitude (closer to 0 is closer). Overlap is $\sum \min(p_T, p_S)$ over the top-50. Gap, overlap and top-50 membership use the teacher at its native temperature ($\tau = 1$). Matching $\tau$ equalizes teacher and student entropy. The last column is $\piz$'s NLL on each teacher's own unfiltered first attempts on the 172 tasks, at its default effort. The lower rows score teachers on other students' own rollouts, thinking tokens only where code is not reported: first-update rollouts of the base-SFT student ($\pibase$ after SFT on distant-teacher demonstrations, \cref{sec:warmup}), 60 rollouts of $\pibase$, and first-update rollouts of $\pisftf$ (\cref{sec:flash,sec:prereg}).}
\label{tab:proximity}
\begin{tabular}{@{}lccccc@{}}
\toprule
Teacher & Gap & Top-50 overlap & In teacher top-50 & Matching $\tau$ & $\piz$ NLL on teacher text \\
\midrule
$\Trl$ (future self) & $-0.021$ / $-0.016$ & 0.952 / 0.963 & 99.90\% / 99.90\% & 0.98 / 0.98 & 0.718 / 0.475 \\
Qwen3.8-27B & $-0.121$ / $-0.092$ & 0.885 / 0.911 & 99.78\% / 99.79\% & 0.89 / 0.87 & 1.291 / 0.095 \\
$\Tf$ (Flash) & $-0.123$ / $-0.099$ & 0.877 / 0.901 & 99.78\% / 99.79\% & 0.79 / 0.78 & 1.708 / 0.145 \\
\midrule
$\Tf$ on base-SFT student & $-0.272$ / $-0.076$ & 0.819 / 0.958 & 99.38\% / 99.93\% & 0.69 / $<$0.6 & --- \\
$\Tf$ on $\pibase$ & $-0.161$ / --- & 0.837 / --- & --- & 0.87 / --- & --- \\
$\Trl$ on base-SFT student & $-0.344$ / --- & 0.805 / --- & --- & --- & --- \\
\midrule
$\Trl$ on $\pisftf$ & $-0.096$ / --- & 0.908 / --- & --- & --- & --- \\
Qwen3.8-27B on $\pisftf$ & $-0.126$ / --- & 0.897 / --- & --- & --- & --- \\
$\Tf$ on $\pisftf$ & $-0.129$ / --- & 0.891 / --- & --- & --- & --- \\
\bottomrule
\end{tabular}
\end{table}

\subsection{Entropy on student prefixes}
\label{sec:entropy}

\begin{figure}[t]
\centering
\begin{tikzpicture}
\begin{axis}[
  name=think, width=0.47\linewidth, height=4.4cm,
  title={\small Thinking tokens}, xlabel={\small teacher $\tau$}, ylabel={\small entropy (nats/token)},
  xmin=0.58, xmax=1.02, ymin=0.25, ymax=0.82, tick label style={font=\scriptsize},
  legend style={at={(0.02,0.98)},anchor=north west,font=\scriptsize,draw=none,fill=none},
]
\addplot[mark=*,blue!70!black] coordinates {(0.6,0.367) (0.7,0.430) (0.8,0.494) (0.9,0.558) (1.0,0.622)};
\addlegendentry{$T_{\mathrm{RL}}$}
\addplot[mark=triangle*,green!45!black] coordinates {(0.6,0.390) (0.7,0.465) (0.8,0.542) (0.9,0.621) (1.0,0.702)};
\addlegendentry{27B}
\addplot[mark=square*,orange!80!black] coordinates {(0.6,0.448) (0.7,0.533) (0.8,0.621) (0.9,0.709) (1.0,0.799)};
\addlegendentry{Flash}
\addplot[dashed,black,domain=0.58:1.02] {0.612};
\addlegendentry{$\pi_0$ (student)}
\end{axis}
\begin{axis}[
  at={(think.east)}, anchor=west, xshift=1.2cm, width=0.47\linewidth, height=4.4cm,
  title={\small Code tokens}, xlabel={\small teacher $\tau$},
  xmin=0.58, xmax=1.02, ymin=0.25, ymax=0.82, tick label style={font=\scriptsize},
]
\addplot[mark=*,blue!70!black] coordinates {(0.6,0.283) (0.7,0.332) (0.8,0.382) (0.9,0.433) (1.0,0.483)};
\addplot[mark=triangle*,green!45!black] coordinates {(0.6,0.309) (0.7,0.370) (0.8,0.432) (0.9,0.495) (1.0,0.560)};
\addplot[mark=square*,orange!80!black] coordinates {(0.6,0.352) (0.7,0.421) (0.8,0.492) (0.9,0.564) (1.0,0.638)};
\addplot[dashed,black,domain=0.58:1.02] {0.476};
\end{axis}
\end{tikzpicture}
\caption{Teacher entropy on $\piz$'s own rollouts as a function of scoring temperature, against the student's entropy (dashed). The future-self teacher crosses the student's line near its native temperature; the distant teachers must be cooled to 0.78--0.89.}
\label{fig:entropy}
\end{figure}
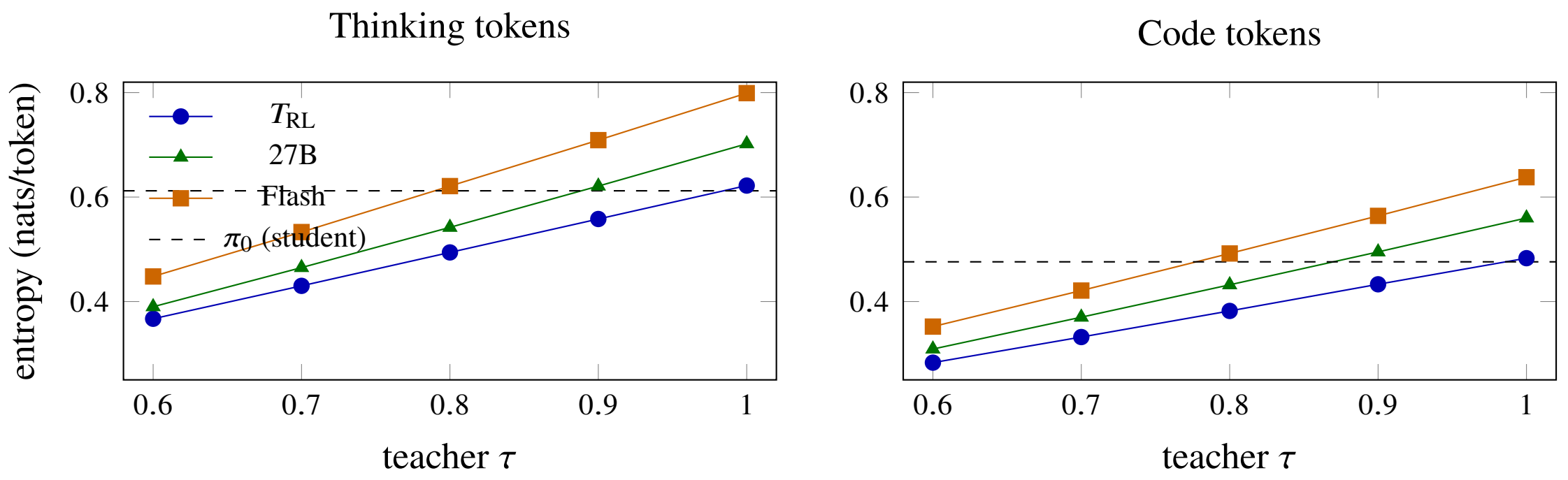

On $\piz$'s rollouts (\cref{fig:entropy}), $\Trl$ matches the student's entropy at $\tau = 0.98$, essentially its native temperature. The 27B must be cooled to 0.87--0.89 and $\Tf$ to 0.78--0.79. At the $\tau = 0.8$ used in training, $\Trl$ was therefore sharper than the student in thinking (0.494 vs.\ 0.612 nats per token), while $\Tf$ was close to it (0.621). On $\pisftf$'s rollouts, $\Tf$ matched at 0.81 in thinking and 0.71 in code, which set the $\tau = 0.8$ used from then on; the first Flash arms scored at $\tau = 1.0$ (\cref{tab:flash}).

Because the future self scored at $\tau = 0.8$ was sharper than the student, part of its gain could come from sharpening alone. Two controls show that sharpening explains only a small part. With the full recipe, $\piz$ taught by its own frozen copy at $\tau = 0.8$ gained 1.77 $\avgk{5}$ (CI $[+0.35, +3.29]$), about a quarter of the future self's $+6.94$, and stayed 5.16 points below it (CI $[-7.17, -3.34]$). With the future self at its entropy-matching $\tau = 0.98$, two seeds reached 12.70 and 13.88, within noise of the $\tau = 0.8$ runs with the same seeds ($-1.33$, CI $[-2.77, +0.13]$; $+0.28$, CI $[-1.03, +1.50]$). The $\tau = 0.98$ runs never developed repeated lines (share at most 0.012 of tokens), while the self-teacher at $\tau = 0.8$ did, with repeated-line share peaking at 0.238 at update 4 and finishing at 0.143.

Entropy and length are separate failure modes. In two controls the Flash-route student $\pisftf$ served as its own frozen teacher. At $\tau = 1$, thinking length did not move (for example $+3.7$K, CI $[-6.8, +13.3]$K at update 3). Flattened to $\tau = 1.15$, the teacher raised student entropy from 0.43 to 0.66 after one update, while thinking moved only $+5.6$K ($[-2.4, +13.6]$K). Calibration therefore addresses entropy drift but cannot by itself explain or fix length runaway.

\subsection{Stopping agreement}
\label{sec:stopping}

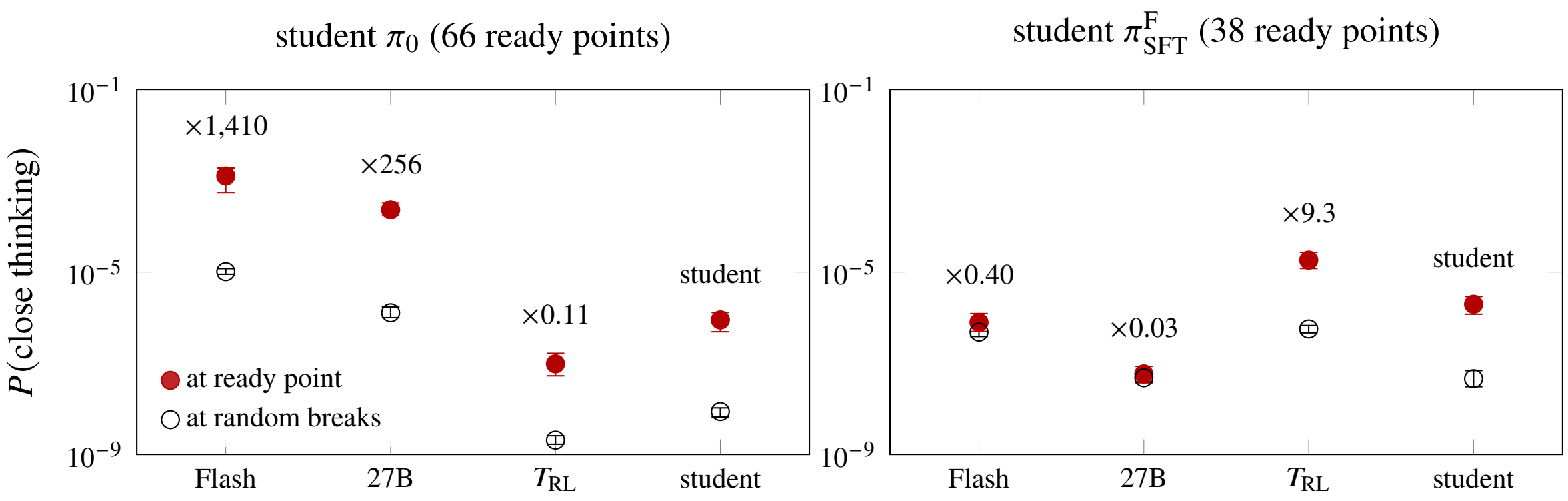
\begin{figure}[t]
\centering
\begin{tikzpicture}
\begin{axis}[ymode=log, width=0.5\linewidth, height=5.2cm, ymin=1e-9, ymax=1e-1, symbolic x coords={F,Q,R,S}, xtick=data, xticklabels={Flash,27B,$T_{\mathrm{RL}}$,student}, enlarge x limits=0.18, tick label style={font=\scriptsize}, label style={font=\small}, error bars/y dir=both, error bars/y explicit, name=p0, title={\small student $\pi_0$ (66 ready points)}, ylabel={\small $P(\text{close thinking})$},
  legend style={at={(0.02,0.02)},anchor=south west,font=\scriptsize,draw=none,fill=white,fill opacity=0.85,text opacity=1}, legend cell align=left]
\addplot[only marks,mark=*,mark size=2.5pt,red!70!black] coordinates {(F,1.260e-03) += (0,6.300e-04) -= (0,7.200e-04) (Q,2.290e-04) += (0,9.500e-05) -= (0,5.400e-05) (R,9.710e-08) += (0,6.790e-08) -= (0,4.410e-08) (S,8.930e-07) += (0,4.070e-07) -= (0,4.020e-07)};
\addlegendentry{at ready point}
\addplot[only marks,mark=o,mark size=2.5pt,black] coordinates {(F,1.020e-05) += (0,1.700e-06) -= (0,1.260e-06) (Q,1.280e-06) += (0,4.300e-07) -= (0,2.800e-07) (R,2.060e-09) += (0,5.100e-10) -= (0,3.900e-10) (S,8.720e-09) += (0,1.780e-09) -= (0,2.090e-09)};
\addlegendentry{at random breaks}
\node[font=\scriptsize,anchor=south] at (axis cs:F,4.16e-03) {$\times$1{,}410};
\node[font=\scriptsize,anchor=south] at (axis cs:Q,7.13e-04) {$\times$256};
\node[font=\scriptsize,anchor=south] at (axis cs:R,3.63e-07) {$\times$0.11};
\node[font=\scriptsize,anchor=south] at (axis cs:S,2.86e-06) {student};
\end{axis}
\begin{axis}[ymode=log, width=0.5\linewidth, height=5.2cm, ymin=1e-9, ymax=1e-1, symbolic x coords={F,Q,R,S}, xtick=data, xticklabels={Flash,27B,$T_{\mathrm{RL}}$,student}, enlarge x limits=0.18, tick label style={font=\scriptsize}, label style={font=\small}, error bars/y dir=both, error bars/y explicit, at={(p0.east)}, anchor=west, xshift=0.8cm, title={\small student $\pi^{\mathrm{F}}_{\mathrm{SFT}}$ (38 ready points)}]
\addplot[only marks,mark=*,mark size=2.5pt,red!70!black] coordinates {(F,7.850e-07) += (0,4.450e-07) -= (0,2.870e-07) (Q,5.840e-08) += (0,2.700e-08) -= (0,2.120e-08) (R,1.830e-05) += (0,8.800e-06) -= (0,6.300e-06) (S,1.970e-06) += (0,9.400e-07) -= (0,7.800e-07)};
\addplot[only marks,mark=o,mark size=2.5pt,black] coordinates {(F,4.820e-07) += (0,2.050e-07) -= (0,9.700e-08) (Q,4.800e-08) += (0,1.290e-08) -= (0,9.100e-09) (R,5.650e-07) += (0,1.090e-07) -= (0,1.010e-07) (S,4.580e-08) += (0,2.330e-08) -= (0,1.520e-08)};
\node[font=\scriptsize,anchor=south] at (axis cs:F,2.71e-06) {$\times$0.40};
\node[font=\scriptsize,anchor=south] at (axis cs:Q,1.88e-07) {$\times$0.03};
\node[font=\scriptsize,anchor=south] at (axis cs:R,5.96e-05) {$\times$9.3};
\node[font=\scriptsize,anchor=south] at (axis cs:S,6.40e-06) {student};
\end{axis}
\end{tikzpicture}
\caption{Probability of closing the thinking block at the student's ready points (filled) and at random paragraph breaks (open), under each teacher and under the student itself; medians with 95\% trace-bootstrap intervals. Labels give the teacher-to-student ratio of the ready-point medians. On $\pi_0$, Flash assigned a median 0.13\% probability to closing at a ready point, about 1,400 times the student's probability. The 27B teacher's ratio was about 260. On $\pi^{\mathrm{F}}_{\mathrm{SFT}}$, Flash barely separates ready points from random breaks, consistent with the Flash route's runaway. The future self stays within an order of magnitude of each student and separates the two positions clearly. A pre-registered test on two further students did not confirm that this pattern predicts length drift (\cref{sec:prereg}).}
\label{fig:stop}
\end{figure}

We asked whether a teacher could tell when the student already had an answer. For each attempt ending in a compiling program, we looked for a point where the student had drafted nearly the same program inside its reasoning. At that \emph{ready point}, we compared the teacher's probability of closing the thinking block on the next token with the student's own probability. We also compared each teacher's closing probability there with its probability at random breaks in the same attempt (\cref{fig:stop}).

This is an indirect measure of the teacher's preference. The training loss masks the close token, so the student's update can reflect that preference only through other sampled tokens. \Cref{app:diagnostics} gives the matching rule, sample counts and uncertainty estimates.

\begin{itemize}[leftmargin=*,itemsep=1pt]
  \item \textbf{Distant teachers on the released student.} At the ready points, the median probability Flash assigned to ending the reasoning block on the next token was 0.13\%. This was about 1,400 times the student's own probability (CI 697--2,460). For the 27B teacher the ratio was about 260 (CI 187--433). Both matched-swap arms then collapsed to short reasoning (\cref{sec:matched}), consistent with this preference.
  \item \textbf{The future self on the released student.} $\Trl$'s closing probability at the ready point is 0.11 times the student's (CI 0.07--0.16), a mild push toward continuing, yet it still rises 47-fold from a random break to the ready point. At the same ready points, the 40-update teacher's closing probability was 0.683 times the student's (CI 0.516--0.976) and 187 times its own probability at random breaks (CI 109--316). Under the final teacher, the student's median thinking stayed between 37K and 63K tokens.
  \item \textbf{Flash on the student trained on its demonstrations.} On the warmed-up student, $\Tf$'s ratio is 0.40 (CI 0.25--0.72), close to the future self's on $\piz$, but $\Tf$ barely distinguishes a ready point from a random break (1.6-fold, CI 0.94--2.69, which includes 1). The 27B behaves the same way on this student (ratio 0.03, contrast 1.2, CI 0.75--1.86). These two types of break had similar closing probabilities under Flash, and the student's reasoning grew to the length cap (\cref{sec:flash}).

  An earlier inspection of 16 traces from the Flash-distilled student agrees: it kept reasoning a median 15K tokens after it had a finished program. Loops make this worse. Repeated lines carry a much smaller teacher--student gap than first-occurrence reasoning, so a long tail of cheap repeats dilutes the per-token penalty. Cutting the loop out of the return also removed the negative return that had reached the decisions before loop onset ($-7.94$ to $-0.37$ on one audited batch), and that variant kept looping.
\end{itemize}
Two properties coincide with the outcomes: the teacher's closing probability relative to the student's, whose large excess coincided with collapse, and the teacher's contrast between ready points and other positions, whose absence coincided with runaway. The close teacher stays within an order of magnitude of the student and has a clear contrast; each distant teacher fails one of the two, depending on the student. These coincidences held for the three pairs above. They did not survive a pre-registered test on two further students (\cref{sec:prereg}).

\paragraph{Other accounts.} The probe does not isolate a mechanism. A simpler account also fits the collapse: the distant teachers' gap is about six times larger, and the return sums it over every later token, so it acts as a penalty that grows with the remaining length (on $\piz$'s rollouts the mean $A_t$ is 70--100 times the mean $d_t$ for all three teachers). A larger gap also gives the teacher term more weight against the reference-KL anchor. \Cref{sec:lengthterm} tests both: neither removing nor centering the return prevents the score loss, and a smaller learning rate only softens it.

\subsection{Starting NLL screens SFT demonstrations}
\label{sec:nll}

Before SFT, we measured the student's negative log likelihood (NLL) on each teacher's demonstrations. Higher NLL means the starting student assigned lower probability to the teacher's text. The 1,704 passing Kimi K3 traces had higher NLL than the 1,247 Flash traces on both reasoning (1.876 vs.\ 0.547 nats per token) and answers (0.188 vs.\ 0.081). A later filter retained 1,110 Flash traces for $\pisftf$.

After one epoch on K3 traces (learning rate $5\times10^{-6}$), the student copied the terse planning but wrote programs that rarely compiled. It scored 0.33 $\avgk{5}$ with 21.9\% compilation, compared with 6.88 and 69.1\% after Flash SFT. Within the K3 run, traces with higher starting NLL also had higher NLL after training (Spearman 0.93). Flash SFT produced occasional strong answers. Its $\bestk{5}$ differed from $\Trl$ by $-1.45$ (CI $[-5.52,+2.60]$), while mean score was 6.88 against 14.65.

Low NLL alone did not guarantee a score gain. The 459 hinted 27B traces were the easiest to predict (0.401 nats per thinking token). After two epochs on them, the student scored 6.28 $\avgk{5}$ against 6.26 for $\piz$ on the same older harness.

\section{Tests of Teacher Transfer on Code}
\label{sec:transfer_tests}

\subsection{Supervised warm-up with distant teachers}
\label{sec:warmup}

Prior work prepares the student with SFT on the same teacher's outputs before OPD \citep{rethinkingopd2026,synergy2026}. We prepared such a student by full SFT from $\pibase$. The first round used 1,905 graded demonstrations on 274 training problems, mostly from the 27B and $\Tf$ (about 105M tokens per pass, equal loss weight on reasoning and answers); four later rounds changed the data mix and schedule.

No checkpoint clearly beat the first round's eighth epoch (6.69 $\avgk{5}$, 92/172 coverage; round 5's best was $-0.31$, CI $[-1.70, +1.17]$), and none matched $\pibase$ itself (7.99). We use that epoch as the starting point for Flash OPD under the same recipe.

On this student's first-update rollouts, $\Tf$ was not closer than it is to $\piz$ where it matters (``$\Tf$ on base-SFT student'' in \cref{tab:proximity}). On thinking tokens the gap was more than twice as large ($-0.272$ vs.\ $-0.123$ on $\piz$), the top-50 overlap lower (0.819 vs.\ 0.877) and the matching temperature 0.69. Only on code was $\Tf$ closer (gap $-0.076$, overlap 0.958). Scoring $\Tf$ on 60 rollouts of $\pibase$ itself isolates the effect of SFT: $\Tf$ is closer to the raw base model than to the SFT student (thinking gap $-0.161$ vs.\ $-0.272$, overlap 0.837 vs.\ 0.819, matching temperature 0.87 vs.\ 0.69). SFT on demonstrations mostly from the 27B and $\Tf$ moved the student \emph{away} from $\Tf$ on its own text.

Flash OPD from this student ran away. Training-time median thinking rose from 33K to 97K tokens over ten updates and forced closes from 12\% to 48\%. On the held-out evaluation it scored 3.61 $\avgk{5}$ and 11.78 $\bestk{5}$, paired $-3.08$ (CI $[-4.58, -1.70]$) and $-5.26$ ($[-9.13, -1.51]$) against its start, with coverage down from 92 to 62 tasks. SFT on the distant teacher's outputs did not make it a usable OPD teacher.

\subsection{Removing or centering the return does not prevent the score loss}
\label{sec:lengthterm}

With $\lambda > 0$ the return of \cref{eq:return} sums the per-token gap over every later token, so a larger gap acts as a penalty that grows with the remaining length. A larger gap also gives the teacher term more weight against the reference-KL anchor; with AdamW the step size itself is set mainly by the learning rate. We tested both accounts in the matched-swap setting (student $\piz$, same recipe and task draws, ten updates, one seed per arm; \cref{tab:lengthterm}). The \emph{centered return}
\begin{equation}
  A_t \;=\; (d_t - \bar d) \;+\; \lambda \sum_{k \ge 1} \gamma^{k} (d_{t+k} - \bar d),
\end{equation}
with $\bar d$ the batch mean of $d_t$ over thinking tokens, removes the term that grows with the remaining length and any uniform push on every token, while keeping the teacher's relative preferences.

\begin{table}[t]
\centering
\small
\caption{Length-term controls from $\piz$ (172 tasks $\times$ 5, ten updates, matched-swap recipe and task draws, one seed per arm). ``Original'' arms are the seed-0 matched-swap arms of \cref{tab:main} ($\lambda = 0.01$). $\Delta$ is paired against $\piz$ (7.10); $^{*}$ marks intervals excluding zero.}
\label{tab:lengthterm}
\setlength{\tabcolsep}{4pt}
\begin{tabular}{@{}llccccc@{}}
\toprule
Teacher & Variant & $\avgk{5}$ & $\bestk{5}$ & Coverage & Median thinking & $\Delta\avgk{5}$ vs.\ $\piz$ \\
\midrule
$\Tf$ & original & 3.07 & 8.72 & 49 & 1.4K & $-4.03^{*}$ \\
 & centered return & 3.53 & 8.53 & 46 & 9.8K & $-3.57$ $[-5.43, -1.92]^{*}$ \\
 & no return ($\lambda = 0$) & 3.15 & 8.36 & 52 & 7.3K & $-3.95$ $[-5.78, -2.34]^{*}$ \\
 & $\lambda = 0.0017$ & 3.98 & 10.16 & 53 & 8.0K & $-3.12$ $[-5.01, -1.36]^{*}$ \\
 & learning rate $/6$ & 4.48 & 11.78 & 58 & 10.7K & $-2.61$ $[-4.18, -1.24]^{*}$ \\
\midrule
Qwen3.8-27B & original & 3.71 & 8.91 & 50 & 3.0K & $-3.39^{*}$ \\
 & centered return & 3.84 & 9.45 & 47 & 33.7K & $-3.26$ $[-4.97, -1.76]^{*}$ \\
 & no return ($\lambda = 0$) & 4.53 & 10.24 & 51 & 51.5K & $-2.56$ $[-4.26, -0.92]^{*}$ \\
\bottomrule
\end{tabular}
\end{table}

\paragraph{Flash.} All variants scored below the released student's 7.10. Centering the return gave 3.53 and removing it gave 3.15, changes of $+0.46$ and $+0.08$ relative to the original Flash arm. Both intervals included zero. In both runs, reasoning shortened about fourfold in the first update. By update four median length was about 4K tokens and entropy about 0.09.

Two changes reduced the loss. We scaled the return weight to $\lambda=0.0017$ so that weight times mean gap matched the future-self run ($0.01\times0.021/0.123$). This gained 0.91 points over the original Flash arm (CI $[+0.11,+1.79]$). A sixfold reduction in learning rate gained 1.41 ($[+0.29,+2.67]$). Both students still finished 2.6--3.1 points below their starting score.

\paragraph{27B.} Centering changed mean score by $+0.14$ relative to the original 27B arm. Removing the return kept reasoning long (median 51K tokens at evaluation, 32\% forced closes), yet the student scored 2.56 points below its starting level. Across 33 runs, the return-based prediction accounted for little of the variation in length change ($R^2=0.014$, \cref{app:diagnostics}).

\subsection{A pre-registered test of the stop probe on two further students}
\label{sec:prereg}

The first three pairs suggested that a teacher's closing probability might predict how long its student would reason. To test that idea, we measured closing probabilities on the first rollouts of $\pisftf$ and the base model after SFT. For four teacher--student pairs we recorded a prediction about reasoning length and the sign of the score change before training. All four runs used the matched-swap LoRA recipe for ten updates. \Cref{tab:prereg} shows the predictions alongside the outcomes. The registered thresholds are in \cref{app:diagnostics}.

\begin{table}[t]
\centering
\footnotesize
\caption{Pre-registered teacher--student grid (LoRA, ten updates, 172 tasks $\times$ 5). Gap, overlap and the stop probe (ratio of ready-point medians, with the teacher's ready-vs-random contrast) were measured on each student's first-update rollouts before training. $\Delta$ is paired against the student; $^{*}$ marks intervals excluding zero. The last row is the Flash run from the base-SFT student above, which was not pre-registered. The base-SFT stop probe rests on 7 rollouts with a ready point. $^{\dagger}$0.68$\times$ the first update's median at update 10, after reaching the cap at update 4.}
\label{tab:prereg}
\setlength{\tabcolsep}{3pt}
\resizebox{\linewidth}{!}{%
\begin{tabular}{@{}lccccccc@{}}
\toprule
Teacher $\to$ student & Gap & Overlap & Stop probe & Predicted & Observed drift & $\Delta\avgk{5}$ & $\Delta\bestk{5}$ \\
\midrule
$\Trl \to \pisftf$ & $-0.096$ & 0.908 & 9.3$\times$ (32$\times$) & shorter, $+$ & runaway, loops & $-1.28$ $[-2.74, +0.28]$ & $-8.78^{*}$ \\
27B $\to \pisftf$ & $-0.126$ & 0.897 & 0.03$\times$ (1.2$\times$) & runaway, $-$ & mild shortening$^{\dagger}$, loops & $-4.18$ $[-5.70, -2.70]^{*}$ & $-13.84^{*}$ \\
$\Tf \to \pisftf$ & $-0.129$ & 0.891 & 0.40$\times$ (1.6$\times$) & runaway, $-$ & runaway & $+0.48$ $[-0.99, +2.12]$ & $-4.50^{*}$ \\
$\Trl \to$ base-SFT & $-0.344$ & 0.805 & 19.7$\times$ (16.5$\times$) & shorter, $-$ & runaway, loops & $-3.27$ $[-4.79, -1.80]^{*}$ & $-7.76^{*}$ \\
\midrule
$\Tf \to$ base-SFT & $-0.272$ & 0.819 & --- & --- & runaway & $-3.08$ $[-4.58, -1.70]^{*}$ & $-5.26^{*}$ \\
\bottomrule
\end{tabular}}
\end{table}

\paragraph{The stop probe failed.} One of four length predictions was right (\cref{tab:prereg}). Both $\Trl$ arms, predicted to shorten, ran away. The 27B arm reached the 98K cap by update 4 and then fell into short looping outputs, ending at 0.68 times its starting length. Only Flash on $\pisftf$ ran away as predicted. Three of the four arms degenerated into repetition loops on lines such as ``OK, I think the code is correct. Let me finalize it.'' In the two such arms on $\pisftf$, training entropy fell to 0.01--0.02 and more than half of the thinking in nearly every rollout consisted of repeated lines; in the base-SFT arm the repeated share reached 0.42 by update 9. Two of four score signs were right.

\paragraph{Teacher choice on two other students.} None of the four runs showed a clear gain in mean score. Across runs, $\bestk{5}$ fell by 4.5--13.8 points. Coverage dropped from 86 to 25--77 tasks on $\pisftf$ and from 92 to 41 on the base-SFT student. For $\pisftf$, the closest teacher was the original student's future self. Mean score fell by 1.28 points under that teacher. Flash had the largest gap and the highest observed change in mean score ($+0.48$). Both intervals against the starting student included zero (\cref{tab:prereg}).

Looking back across students, every teacher in this grid had a reasoning-token gap at least 4.6 times as large in magnitude as the successful future-self pairing on $\piz$ ($-0.096$ or more negative, compared with $-0.021$). This comparison was retrospective. It leaves open how close a new teacher must be to help.

\section{Replication on a Public Math Pair}
\label{sec:replication}

We tested the code result on competition math with a fresh student and two public teachers. DeepScaleR-1.5B-Preview is the future self of the starting DeepSeek-R1-Distill-Qwen-1.5B student. It was trained from that checkpoint with GRPO under increasing context lengths \citep{deepscaler2025}. The other teacher, DeepSeek-R1-Distill-Qwen-32B, was distilled separately from R1 \citep{guo2025deepseekr1}. It was the stronger solver on our evaluation (77.4\% vs.\ 57.0\%). The OPD prompts came from DeepScaleR's training set, so the future-self teacher had prior RL exposure to those problems.

On the starting student's reasoning tokens, the future self had a smaller gap ($-0.022$ vs.\ $-0.067$ nats per token) and more top-50 overlap (0.943 vs.\ 0.922). We predicted it would teach the student more effectively. The three pre-registered arms used the future self, the 32B teacher, and the 32B teacher with the centered return of \cref{sec:lengthterm}. We ran two seeds per arm for 50 updates.

We evaluated each checkpoint on 30 AIME 2024, 30 AIME 2025 and 83 AMC~12 problems. Each of the 143 problems contributed equally to accuracy, averaged over 16 attempts per problem with a 32,768-token budget (\cref{tab:replication}). \Cref{app:math-eval} documents sampling, dataset checks and comparisons with published scores.

\begin{table}[t]
\centering
\small
\caption{Public math replication (avg@16 accuracy, 32K budget). Top: baselines per evaluation set; ``pooled'' weights the 143 problems equally. Bottom: pooled accuracy after 50 OPD updates, per seed and averaged; $\Delta$ is the seed mean paired against the student; median length is relative to the student's 10.3K tokens; $^{*}$ marks intervals excluding zero.}
\label{tab:replication}
\begin{tabular}{@{}lcccc@{}}
\toprule
Model & AIME 24 & AIME 25 & AMC & Pooled \\
\midrule
Student (R1-Distill-1.5B) & 0.279 & 0.237 & 0.630 & 0.474 \\
Future self (DeepScaleR-1.5B) & 0.402 & 0.287 & 0.733 & 0.570 \\
Distant (R1-Distill-32B) & 0.717 & 0.523 & 0.885 & 0.774 \\
\bottomrule
\end{tabular}

\vspace{6pt}
\begin{tabular}{@{}lccccc@{}}
\toprule
OPD teacher & Seed 0 & Seed 1 & Mean & $\Delta$ vs.\ student & Median length \\
\midrule
Future self & 0.549 & 0.556 & \textbf{0.552} & $+0.079$ $[+0.053, +0.105]^{*}$ & 0.57$\times$ \\
Distant & 0.479 & 0.486 & 0.482 & $+0.009$ $[-0.009, +0.026]$ & 1.00$\times$ \\
Distant + centered return & 0.302 & 0.375 & 0.339 & $-0.135$ $[-0.167, -0.106]^{*}$ & 3.20$\times$ (cap) \\
\bottomrule
\end{tabular}
\end{table}

\paragraph{Result.} OPD from the future self raised the student from 0.474 to 0.552 (two-seed mean, $+0.079$), 82\% of the way to its teacher, while shortening its reasoning to 0.57 times the student's median length. OPD from the stronger distant teacher changed accuracy by $+0.009$ (CI $[-0.009,+0.026]$), and the future self beat it by 0.070 (CI $[+0.049, +0.092]$).

With the centered return, median length reached the 32K cap. Accuracy fell 14.4 percentage points relative to the plain 32B arm (95\% CI for the loss $[11.5,17.4]$) and 13.5 points relative to the starting student ($[10.6,16.7]$).

\paragraph{Robustness across benchmarks.} The 83 AMC problems account for 58\% of the registered pool. On the 60 AIME problems alone, the future-self advantage was 7.1 percentage points (95\% CI $[4.1,10.5]$). The ordering held on each AIME year and each AMC year (\cref{tab:math_subsets}). In each seed separately, every reported subset also had a positive lower confidence bound. These subset intervals are descriptive, with no adjustment for multiple comparisons.

\begin{table}[htbp]
\centering
\small
\caption{Math accuracy by problem set, averaged over two training seeds (percent). $\Delta$ is the future-self arm minus the plain 32B arm in percentage points, with paired 95\% problem-bootstrap intervals. The last row is the pre-registered comparison. The subset rows overlap.}
\label{tab:math_subsets}
\begin{tabular}{@{}lrcccc@{}}
\toprule
Set & Problems & Future self & 32B & 32B centered & $\Delta$ [95\% CI] \\
\midrule
AIME 2024 & 30 & 38.0 & 29.6 & 19.3 & $+8.4$ $[4.0,14.1]$ \\
AIME 2025 & 30 & 28.7 & 22.9 & 18.1 & $+5.7$ $[2.2,9.8]$ \\
AIME combined & 60 & 33.3 & 26.3 & 18.7 & $+7.1$ $[4.1,10.5]$ \\
AMC 2022 & 43 & 64.1 & 57.4 & 39.7 & $+6.7$ $[3.0,10.9]$ \\
AMC 2023 & 40 & 78.6 & 71.3 & 50.4 & $+7.3$ $[3.0,12.1]$ \\
AMC combined & 83 & 71.1 & 64.1 & 44.8 & $+7.0$ $[4.1,10.0]$ \\
\midrule
Registered pool & 143 & 55.2 & 48.2 & 33.9 & $+7.0$ $[4.9,9.2]$ \\
\bottomrule
\end{tabular}
\end{table}

\paragraph{Pre-registered calls.} Three of eight held: the positive sign for the future self, the negative sign for the distant teacher with the centered return, and the future self beating the distant teacher. The distant arm's interval included both a small loss and a gain. All three length predictions were wrong. We predicted runaway for every arm, with less runaway for the centered arm than for the plain distant arm; instead the future self shortened reasoning, the plain distant arm left it unchanged, and the centered arm ran further than the plain arm. We had predicted that centering would leave accuracy unchanged. As on code, the ordering of scores followed proximity (future self above the distant teacher by 0.070), while the length predictions failed.

\section{Discussion}
\label{sec:discussion}

\paragraph{What the swaps establish.} For both starting students, the teacher with the highest standalone score produced a worse OPD student than the student's own RL-trained descendant. On code this comparison held in both distant-teacher seeds under the same ten-update recipe. On math we predicted that the descendant would produce the better student. Its measured advantage was 7.0 percentage points (\cref{sec:replication}). The starting students, training budgets and score scales differ across domains, so we compare teacher choices within each domain.

\paragraph{Choosing a teacher before training.} On the initial code student's reasoning tokens, the smallest gap picked the future self among the three stronger candidates. The math replication predicted in advance that the future self would yield higher student accuracy than the 32B teacher. For the code student trained on Flash demonstrations, the original student's future self still had the smallest gap. Flash produced the highest observed mean score there, and neither teacher showed a clear gain over the student's starting score (\cref{sec:prereg}). The starting student itself has zero gap at native temperature, yet using it as a sharpened teacher gained only 1.77 code points in seed~0 (\cref{sec:futureself}).

Both winners were descendants, and the math teacher had trained on the OPD prompt pool. The 40-update code teacher is closer to its starting model than the 124-update teacher (\cref{sec:proximity}), which gives a test within one lineage. To test direct descent, we would compare that lineage with a separately trained teacher matched on standalone score and the initial reasoning-token gap. A numerical forecast of student gain would require calibration on additional pairs and evaluation on held-out students and teacher lineages.

\paragraph{What the Flash comparisons add.} Flash solved more of the code benchmark itself, yet scoring the released student's own tokens with Flash lowered its score within two updates. Starting with Flash demonstrations, changing the token loss and adding program execution rewards each tested a different explanation for that gap. Matching entropy still left many completions at the length cap. An outcome-dominated continuation scored above the SFT-only student, but its mean gain over the checkpoint where mixing began was uncertain and its best-of-five score fell (\cref{sec:mixed}). A teacher's own solutions, its scores on student reasoning and the grader's assessment of the finished program supply different information. Across the recipes tested here, none made Flash as effective for this student as its future self.

\paragraph{When future-self OPD is useful.} Given an existing RL teacher, the seed-0 student gained 3.86 more $\avgk{5}$ points than direct RL from the same initialization at its nearest saved token budget (\cref{sec:budget}). Building the teacher first used about 747M trainer tokens. In these experiments distillation took the student back to within 0.6--1.7 points of a model already available. A practical use would be to transfer those gains to a sibling initialization. In our one such test the change from $\pisftf$ was $-1.28$ in $\avgk{5}$ (CI includes zero) and $-8.78$ in $\bestk{5}$ (\cref{sec:prereg}). The successful transfers to the teacher's own initialization are consistent with an iterate-ahead, distill-back loop \citep{trainahead2026}.

\paragraph{Evaluation practice.} Mean and best-of-five scores answer different questions. A change that improved average partial credit could still reduce the best program found in five attempts. Reasoning length also depended on the task draw. The initial median for $\piz$ ranged from 50K to 63K tokens across the draws in \cref{fig:dynamics}. We therefore compared length on the same tasks and checked that the trainer and rollout server agreed after every weight reload (\cref{sec:checks}).

\section{Limitations}
\label{sec:limitations}

\paragraph{Teacher choice.} The matched swaps cover one released 9B code student and one 1.5B math student. Both winning teachers descended from the corresponding student, and the math teacher had already trained on the distillation prompts. On a different code student the same future-self teacher showed no clear benefit. Calibrating the initial gap as a predictor requires additional students and independently trained teachers.

\paragraph{Training recipe and budget.} Each matched swap held the recipe fixed. On code, both distant teachers reduced score by the second update. A sixfold reduction in the Flash learning rate recovered 1.41 points relative to the original Flash arm, leaving the student below its starting score. These checkpoints cover a small part of the possible schedules. The full-weight Flash experiments changed the starting student and optimizer settings. The 20-update hint pilot had no matched plain-RL control. For standalone teacher scores, the distant models used their own prompt protocols. For the historical RL curve, only the 15-, 40- and 124-update checkpoints plotted as open diamonds were evaluated with the frozen harness (\cref{fig:budget}). Early RL token counts are estimates.

\paragraph{Training variability.} We have three seeds each for future-self code OPD and full-weight Flash OPD. Resampling tasks and seeds preserves the positive future-self gain and leaves the Flash gain unresolved. The Flash serving configuration changed after seed~0. The distant matched swaps, entropy-matched future self and math arms have two seeds each. Most other controls have one, so their intervals describe variation across problems for the observed checkpoints. Three seeds still give limited information about variation across training runs.

\paragraph{Mechanism.} Neither removing nor centering the return prevented the loss of code score. Sharpening the starting student as its own teacher recovered about a quarter of the seed-0 future-self gain. The stop probe predicted length change in one of four pre-registered runs. It rested on 66 and 38 ready points for the two main code students and only seven rollouts with a ready point for the base-SFT student. Three of the four pre-registered runs developed repetition loops whose cause remains unclear (\cref{sec:lengthterm,sec:prereg}).

\paragraph{Reproducibility and artifact availability.} There is currently no planned public release of our custom training or evaluation code, trained checkpoints, task manifests, sync-check implementation or per-attempt outputs. The FrontierCS benchmark and math models are publicly available. \Cref{sec:objective} defines the training objective. The appendix gives the code evaluation and uncertainty rules (\cref{app:code-eval}), training settings and consistency checks (\cref{app:training}), diagnostic estimators (\cref{app:diagnostics}) and math checkpoint revisions and evaluation protocol (\cref{app:math-eval}). The paper reports individual training seeds for the matched swaps. Recomputing paired intervals requires the underlying per-problem scores.

\paragraph{AI assistance.} The author used Claude and Codex to help draft and edit the manuscript and check numerical claims against experiment records.

\bibliographystyle{plainnat}
\bibliography{refs}

\appendix
\section{Code evaluation protocol}
\label{app:code-eval}

\paragraph{Benchmark version.} We used the 172 algorithmic problems (98 standard, 74 interactive) in FrontierCS v1.0.0 at commit \texttt{55c104be}. The first public release described in the FrontierCS paper contained 107 algorithmic problems. Our revision excludes 16 later additions, ten of them synthetic problems of the kind in our RL training pool. Our full-test scores and public-test scores use different test sets.

\paragraph{Prompt, execution and scoring.} We passed the problem statement through the student's stock chat template without a system turn. The sandbox compiled each program with \texttt{g++ -O2} and ran it under the problem's time limit. The score was 100 times the mean per-test-case score ratio. At a forced close the harness inserted a fixed transition sentence and closed the thinking block. We masked the injected sentence from the training loss.

\paragraph{Teacher baselines.} Flash and the 27B used their own samplers and chat templates at their default xhigh reasoning effort, which adds a system turn. The 27B was evaluated on an earlier harness version. On that version $\piz$ scored 6.26, compared with 7.10 on the frozen harness (paired difference $-0.84$, 95\% CI $[-2.30,+0.54]$). The frozen harness was used for the student comparisons in the matched swap.

\paragraph{Uncertainty.} Unless otherwise specified, we resampled problems 10,000 times, using the same problem draw on both sides of a comparison and keeping each problem's five attempts together. For the two routes with three training seeds we also resampled whole seeds across each common problem draw. We call a difference \emph{within noise} when its paired interval includes zero. These intervals are typically $\pm$1.3--2 points wide and can include meaningful gains or losses. Neighboring RL checkpoints differed by about $\pm1$ $\avgk{5}$ point.

\paragraph{Missing grades and grader load.} Missing grades counted as zero. In seed~1 of the two matched distant-teacher runs, all five attempts on task 143 lacked grades in each arm. One further grade on task 164 was missing for Flash, giving six missing grades out of 860 for that checkpoint. The seed-0 Flash update-two evaluation had one additional missing grade on task 249. Grading concurrency ranged from two to about 40 programs. Regrading four checkpoints at concurrency two changed $\avgk{5}$ by at most 0.01.

\subsection{Paired comparisons across code seeds}
\label{app:code-contrasts}

\paragraph{Matched teacher swaps.} The two Flash seeds changed $\avgk{5}$ by $-4.03$ and $-3.69$ relative to $\piz$ (95\% CI $[-5.73,-1.89]$ for seed~1). The two 27B seeds changed it by $-3.39$ and $-3.45$ (CI $[-5.32,-1.81]$ for seed~1). Each distant-teacher checkpoint finished 10.3--11.0 points below future-self seed~0, with intervals excluding zero. The within-teacher difference between seeds was $+0.33$ for Flash and $-0.06$ for 27B. Both intervals included zero. In seed~0, Flash minus 27B was $-0.63$ (CI $[-1.88,+0.48]$).

\paragraph{Future-self transfer.} The three-seed mean gain over $\piz$ had a between-seed standard deviation of 0.52. Its task-only interval was $[+4.68,+8.37]$. The task-only interval for the difference against the teacher was $[-2.23,-0.04]$. Individually, the three students differed from the teacher by $-0.62$ (CI $[-2.01,+0.71]$), $-1.04$ ($[-2.40,+0.18]$) and $-1.65$ ($[-3.03,-0.36]$). \Cref{sec:futureself} gives intervals that also resample seeds.

\paragraph{Full-weight Flash OPD.} The task-only interval for the three-seed mean gain over $\pisftf$ was $[-0.37,+2.16]$. Seeds~1 and~2 differed from the starting student by $-0.05$ (CI $[-1.58,+1.53]$) and $+0.59$ (CI $[-0.83,+2.21]$). Seed~1 scored 2.02 points below seed~0 (CI for the difference $[-3.33,-0.78]$). The change in teacher serving configuration between seed~0 and the other two seeds coincided with the change in training task draws. Its independent contribution cannot be estimated from these runs.

\section{Code training and implementation}
\label{app:training}

\subsection{Training schedules and compute}
\label{sec:compute}

\paragraph{Future-self teacher.} $\Trl$ starts from $\piz$ and receives 124 GRPO updates with learning rate $10^{-6}$, KL coefficient $10^{-3}$ to the initialization and clip ratio 0.2. The first 40 updates use 200 synthetic open-ended problems derived from competitive programming at reduced budgets (57,344 thinking and 8,192 answer tokens), with 8 problems $\times$ 8 samples per update. The remaining 84 use 137 problems (133 of the same kind and 4 AtCoder Heuristic Contest tasks) at the full budgets, keeping 16 of 24 sampled groups of 8 attempts. The reward is the grader's score. During updates 41--109 an LLM judge also re-ranked programs within groups, carrying up to 22\% of the advantage magnitude. After that, a small bonus rewarded programs that beat the task's baseline program.

\paragraph{Training-set overlap.} We screened every training problem against every FrontierCS problem by identifier, title, test hashes, TF-IDF and $n$-gram similarity, and an LLM review. None matched (maximum TF-IDF cosine 0.146).

\paragraph{Matched teacher swaps.} Each update used eight tasks and eight student samples per task. We trained rank-64 LoRA adapters on all projections (173M parameters) with AdamW, a constant learning rate of $4\times10^{-5}$, weight decay 0.01 and gradient clipping at 1.0. Teachers scored the student's sampled tokens using top-50 log probabilities at $\tau=0.8$. Both distant teachers used medium effort. The return used $\gamma=0.9999$ and $\lambda=0.01$, with zero outcome weight and reference KL $\beta=10^{-3}$ to $\piz$. We used the original future-self recipe, including the temperature inherited from an earlier Flash calibration.

\paragraph{Flash demonstration and OPD stages.} We trained $\pisftf$ on 1,110 Flash traces from 268 problems for three epochs (207 steps of 16 traces, learning rate $10^{-5}$ with cosine decay). The five-update experiments in \cref{tab:flash} trained all weights at learning rate $10^{-6}$ and $\beta=10^{-3}$, with eight tasks and eight attempts per task. Flash scored the samples at medium reasoning effort.

\paragraph{Accounting for compute.} Trainer tokens count tokens in sequences processed by the student's training pass. GPU-hours are elapsed wall-clock time multiplied by allocated GPUs, excluding queue time. Of the roughly 747M trainer tokens used to build the RL teacher, 636.2M were logged over the final 84 updates and 111.1M were estimated over the first 40. One trainer token, one decoded token and one token scored by a larger teacher have different costs. Of the approximately 57 GPU-hours allocated to the seed-0 future-self OPD run, 10.3 trainer GPU-hours were measured. The rollout and teacher shares are estimates.

\subsection{Loss implementation}
\label{app:loss}

The code student sampled at temperature 1.0 over the full vocabulary. To apply a scoring temperature to the teacher, we renormalized its top-50 log probabilities and estimated the remaining tail. Against exact full-vocabulary normalization over 300 positions, the mean absolute error was 0.0015 nats (maximum 0.023).

We clipped the teacher term to $\pm5$ per token. Special targets, including the tokens that open and close the thinking block, received neither a teacher term nor a return. The reference KL still covered them. Each token's advantage entered the loss with constant weight $1/131{,}072$, the reciprocal of the sequence budget. We applied no per-batch or per-sequence advantage normalization. The per-token reference-KL estimate was $k_3=r-1-\log r$, with $r=\pi_{\mathrm{ref}}/\pi_\theta$ evaluated on the sampled token.

\subsection{Implementation checks}
\label{sec:checks}

The signal in \cref{eq:dt} is a difference of two log-probabilities of the same token, and it is small: the mean teacher--student gap is 0.02--0.12 nats per token at the teacher's native temperature (\cref{sec:proximity}), and the mean $|d_t|$ under the future self at $\tau = 0.8$ is 0.136 at the first update. Errors of a few hundredths of a nat therefore matter. We enforce three invariants.
\begin{itemize}[leftmargin=*,itemsep=1pt]
  \item \textbf{(C1) Sampling consistency.} If the rollout server samples from a truncated or penalized distribution $q_\theta$ and the student term is read from it, the signal gains the teacher-independent term $\log \pi_\theta - \log q_\theta$. Our code runs sample at temperature 1.0 over the full vocabulary, so $q_\theta = \pi_\theta$.
  \item \textbf{(C2) Optimizer state.} A step with learning rate zero is not a no-op for Adam: the moments still absorb the gradient, and the first moment enters every later update. A dry-run step must skip the optimizer or reset its state.
  \item \textbf{(C3) Weight consistency.} After rounding and adapter merging, the trainer and the rollout server must evaluate the same weights. The trainer's forward pass uses the merged, rounded weights the server uses, while keeping the adapter's gradients. We compare trainer and server log-probabilities on the same sampled tokens before and after every weight reload; in the future-self run the mean absolute difference after reload was 0.0048 nats, down from 0.0125 before this change.
\end{itemize}
Teacher scores were reproducible on a fixed server configuration to about 0.01 nats per token. Changing the teacher server's parallelism and prefix-cache settings moved per-token log-probabilities by 0.022 nats on average, with no shift in the mean signal ($3\times10^{-5}$ nats). The sign of the teacher signal at the thinking-close token says little, since its magnitude there is tiny, so we measure the teacher's closing probability directly (\cref{sec:stopping}).

We also audited the full-weight Flash runs for token alignment, loss signs, masks and aggregation. These checks found no implementation error.

\subsection{Additional Flash length controls}
\label{app:flash-controls}

We compared median thinking length on the same tasks as $\pisftf$. With forward KL on Flash's top 20 tokens the paired change was $+23.0$K by update five. Masking and unmasking special tokens gave changes of $+36.5$K and $+22.0$K. Supervising the close token gave $+44.5$K, RL-style length rules gave $+15.0$K by update two, and weighting the teacher term on thinking tokens by 0.1 gave $+17.0$K. Each interval excluded zero. With reference-KL coefficient 0.5, median thinking still rose from 33.8K to 66.8K by update three.

\section{Diagnostic measurements}
\label{app:diagnostics}

\paragraph{Shared rollouts and KL estimates.} For the initial code-student comparisons in \cref{sec:futureself,sec:proximity}, the 64 unique first-update rollouts comprised 3,888,374 reasoning positions and 738,530 code positions. Both analyses excluded three byte-identical preflight replays. The pure Flash and pure future-self endpoints of the offline teacher mixture reproduced the original probabilities across all 4.63M positions.

For the KL estimates we scored each model at native temperature. We computed a token-weighted top-50 lower bound with the remaining mass pooled into a tail bucket, and resampled whole rollouts 10,000 times. We repeated the calculation on 64 fresh rollouts from the trained student on the same task draw, using a different scoring server. Those rollouts contained 189K code positions, compared with 739K on the original draw.

\begin{center}
\begin{tabular}{@{}lcc@{}}
\toprule
KL direction (nats/token) & Original prefixes & Trained student's prefixes \\
\midrule
Starting student $\|$ teacher & 0.0200 & 0.0245 \\
Trained student $\|$ teacher & 0.0193 & 0.0194 \\
Teacher $\|$ trained student & 0.0258 & 0.0257 \\
Trained student $\|$ starting student & 0.0338 & 0.0333 \\
\bottomrule
\end{tabular}
\end{center}

For the trained student relative to the teacher, the 95\% intervals were $[0.0182,0.0205]$ on the original prefixes and $[0.0189,0.0200]$ on the fresh prefixes.

\paragraph{Matched length comparison.} For the update-five Flash comparison in \cref{sec:futureself}, we generated four fresh attempts from $\pisftf$ on each of the eight training tasks in that batch. We took the difference in mean thinking length for each task and averaged those eight paired differences.

\paragraph{Stopping probe.} We considered student attempts whose final program compiled under the grader's flags. The ready point was the first paragraph break after a code block in the reasoning with line-level similarity of at least 0.8 to the final program. Each trace contributed five random comparison breaks. We compared medians and bootstrapped whole traces 10,000 times. There were 66 ready points in 723 compiling $\piz$ traces and 38 in 154 compiling $\pisftf$ traces. Two versions of the $\piz$ prompt contributed 40 and 26 ready points. Their estimates agreed in direction and magnitude.

At a ready point the student receives no direct teacher term for the masked close token. Under Flash the change in total log probability available to continue was about $\log(1-1.3\times10^{-3})\approx-0.0013$ nats, roughly a hundredth of the mean gap.

\paragraph{Registered predictions for reasoning length.} We predicted shortening when the teacher distinguished ready points from random breaks. We predicted mild shortening when its closing probability was within a factor of ten of the student's. We also predicted shortening for the one pair with a 19.7-fold ratio and a clear contrast. When the teacher showed no contrast, we predicted runaway reasoning.

We classified a run as \emph{runaway} if its update-ten median was at least 1.25 times its first-update median or if forced closes rose by ten percentage points. The threshold for \emph{collapse} was a length ratio at or below 0.5. Ratios above 0.5 and up to 0.95 counted as \emph{mild shortening}.

\paragraph{Length change across runs.} For 33 OPD runs with the return enabled, we regressed the per-update change in median thinking length on the length penalty predicted by the return. The predictor was proportional to learning rate, $\lambda$, the gap and discounted remaining length. It explained little variation overall ($R^2=0.014$) and predicted the correct sign in 42\% of updates. Within the runs from $\piz$ whose reasoning collapsed, the correlation was 0.95.

\section{Math replication protocol}
\label{app:math-eval}

\paragraph{Models and launches.} The pinned revisions were \texttt{ad9f0ae0} for the 1.5B student, \texttt{e3f524ce} for DeepScaleR and \texttt{711ad2ea} for the 32B teacher. They share token IDs. Both teachers scored responses using the student's chat template. DeepScaleR's native template omits the forced opening thinking tag. The pre-registration specified \texttt{p5:1}. All six production launches recorded \texttt{p5.48xlarge:4}.

\paragraph{Problem sets and uncertainty.} The AMC~12 pool consisted of 43 problems from 2022 and 40 from 2023. Repeated source IDs were contest numbers reused across years. We sampled at temperature 0.6 and top-$p$ 0.95. All nine evaluation packets matched the pinned datasets by stable problem key, with no missing, duplicate or reordered problems. We drew 10,000 paired bootstrap samples of problems, keeping each problem's 16 attempts and the two training seeds together. Pairing by problem key reproduced the original row-order estimates.

\paragraph{Overlap with training.} No evaluation problem reached 0.8 containment of its character 5-grams in a training prompt. The optional 13-word $n$-gram filter was disabled in the production data build.

\paragraph{Comparison with published evaluations.} The DeepScaleR authors report 28.8\% and 43.1\% on AIME 2024 and 62.9\% and 73.6\% in a column labelled \emph{AMC 2023} for the student and its future self \citep{deepscaler2025}. DeepSeek reports 72.6\% on AIME 2024 for the 32B model \citep{guo2025deepseekr1}. Our AIME 2024 scores are within three percentage points of these reports. All 83 of our AMC problems, their answers and their order match DeepScaleR's released AMC test file. The evaluation script underlying the card's column was unavailable, so its exact scoring population remains unconfirmed.

\end{document}